\documentclass[final,3p,times,authoryear]{elsarticle}

\usepackage{amsmath,amssymb,amsfonts}
\usepackage{algorithmic}
\usepackage{graphicx}
\usepackage{textcomp}
\usepackage{xcolor}

\usepackage{amsthm}
\usepackage{multirow}
\usepackage{array}
\usepackage{bm}
\usepackage{enumerate}

\usepackage{booktabs}
\usepackage{adjustbox}
\usepackage{relsize}

\usepackage{algorithm}

\usepackage{url}
\usepackage{bbm}
\usepackage{dsfont}
\usepackage{array}
\usepackage{courier}
\usepackage{bbding}
\usepackage{pifont}

\usepackage[colorlinks]{hyperref}

\usepackage{float}  
\usepackage{subfigure} 
\usepackage{makecell}

\journal{Information Processing and Management
}

\begin{document}

\begin{frontmatter}



\title{RoSAS: Deep Semi-supervised Anomaly Detection with Contamination-resilient Continuous Supervision}


\author[address1,address2]{Hongzuo Xu}
\ead{xuhongzuo13@nudt.edu.cn}

\author[address1,address2]{Yijie Wang\corref{mycorrespondingauthor}}
\cortext[mycorrespondingauthor]{Corresponding author}
\ead{wangyijie@nudt.edu.cn}

\author[address3]{Guansong Pang}
\ead{gspang@smu.edu.sg}

\author[address2]{Songlei Jian}
\ead{jiansonglei@nudt.edu.cn}

\author[address4]{Ning Liu}
\ead{liuning17a@nudt.edu.cn}

\author[address2]{Yongjun Wang}
\ead{wangyongjun@nudt.edu.cn}

\address[address1]{
	National Key Laboratory of Parallel and Distributed Computing
}
\address[address2]{College of Computer, National University of Defense  Technology, Changsha Hunan 410073, China
}
\address[address3]{School of Computing and Information Systems, Singapore Management University, Singapore 178902, Singapore
}
\address[address4]{College of Information and Communications, National University of Defense  Technology, Wuhan Hubei 430010, China
}

\begin{abstract}

	Semi-supervised anomaly detection methods leverage a few anomaly examples to yield drastically improved performance compared to unsupervised models. However, they still suffer from two limitations: 1) unlabeled anomalies (i.e., anomaly contamination) may mislead the learning process when all the unlabeled data are employed as inliers for model training; 2) only discrete supervision information (such as binary or ordinal data labels) is exploited, which leads to suboptimal learning of anomaly scores that essentially take on a continuous distribution. Therefore, this paper proposes a novel semi-supervised anomaly detection method, which devises \textit{contamination-resilient continuous supervisory signals}. Specifically, we propose a mass interpolation method to diffuse the abnormality of labeled anomalies, thereby creating new data samples labeled with continuous abnormal degrees. Meanwhile, the contaminated area can be covered by new data samples generated via combinations of data with correct labels. A feature learning-based objective is added to serve as an optimization constraint to regularize the network and further enhance the robustness w.r.t. anomaly contamination. Extensive experiments on 11 real-world datasets show that our approach significantly outperforms state-of-the-art competitors by 20\%-30\% in AUC-PR and obtains more robust and superior performance in settings with different anomaly contamination levels and varying numbers of labeled anomalies. The source code is available at \url{https://github.com/xuhongzuo/rosas/}.
\end{abstract}


\end{frontmatter}

\section{Introduction}

Anomaly detection is to identify exceptional data objects that are deviated significantly from the majority of data, which has wide applications in many vital domains, e.g., network security, financial surveillance, risk management, and AI medical diagnostics \citep{pang2021review}. Anomaly detection is often posited as an unsupervised problem due to the difficulty of accessing adequate labeled data \citep{han2022adbench,jiang2023weakly}. The past decade has witnessed a plethora of unsupervised anomaly detection methods that estimate/learn data normality via various data characteristics (e.g., proximity, probability, 
or clustering membership) or deep models (e.g., different kinds of Autoencoders or generative adversarial networks). However, these unsupervised methods often have many false alarms which can overwhelm human analysts, leading to the failure of investigating real threats. It is challenging, if not impossible, to accurately detect true anomalies of real interest without any prior information indicating what kind of data are anomalies.

In fact, in many real-world applications, there are often a few readily accessible anomaly examples. For example, some abnormal events such as credit card frauds or insiders’ unauthorized access are reported (by users) or logged (in the system). Small genuine anomaly data can be directly retrieved from these records, without requiring extra annotations. 
This naturally inspires us to harness these true anomalies in combination with unlabeled data when training detection models. This learning paradigm falls into the category of semi-supervised learning \citep{chen2019semisupervised,kang2021semi,van2020survey,yu2018adaptive} that permits using small labeled data as well as a large amount of unlabeled data. 
Recently, with the help of dozens of anomaly examples, semi-supervised methods have shown drastically improved detection performance compared to unsupervised methods that work on unlabeled data only \citep{ding2021few,ding2022catching,jiang2023weakly,pang2018learning,pang2019deep,pang2019prenet,zhou2021feature,zhou2022unseen}.

By summarizing prior arts, this paper first proposes a general deep semi-supervised anomaly detection framework by introducing a two-stage network structure and a general learning objective. This framework presents a unifying view of this research line.
More importantly, this framework reveals the following two key limitations of existing deep semi-supervised anomaly detection models that we aim to address in this study:

\textit{Robustness w.r.t. anomaly contamination.}
Many studies \citep{pang2018learning,ruff2020deep,wu2021surrogate,zhou2021feature} assume all the unlabeled data as normal since anomalies are rare events. However, some anomalies are still hidden in the unlabeled set (i.e., \textit{anomaly contamination}). This contamination might disturb anomaly detection models and blur the boundaries of normal patterns, leading to the potential overfitting problem. 
Some attempts \citep{pang2019deep,pang2021explainable,pang2019prenet} have been made to address this problem by using a Gaussian prior when defining optimization targets or using concatenated data pairs as augmented training data.

\textit{Continuous supervision of anomaly score optimization.}
Anomaly detection models are typically required to output anomaly scores to indicate the degree of being abnormal for human investigation of the top-ranked anomalies. 
However, current models only use discrete supervision information, e.g., binary optimization targets \citep{pang2018learning,pang2019deep,pang2021explainable,ruff2020deep,wu2021surrogate,zhou2021feature} or ordinal class labels \citep{pang2020self,pang2019prenet}, to optimize anomaly scores that essentially take on a continuous distribution. The lack of continuous supervision may result in suboptimal learning of anomaly scores. To the best of our knowledge, we are the first to raise this problem in anomaly detection.

\begin{figure}[t]
\centering	\includegraphics[width=0.98\textwidth]{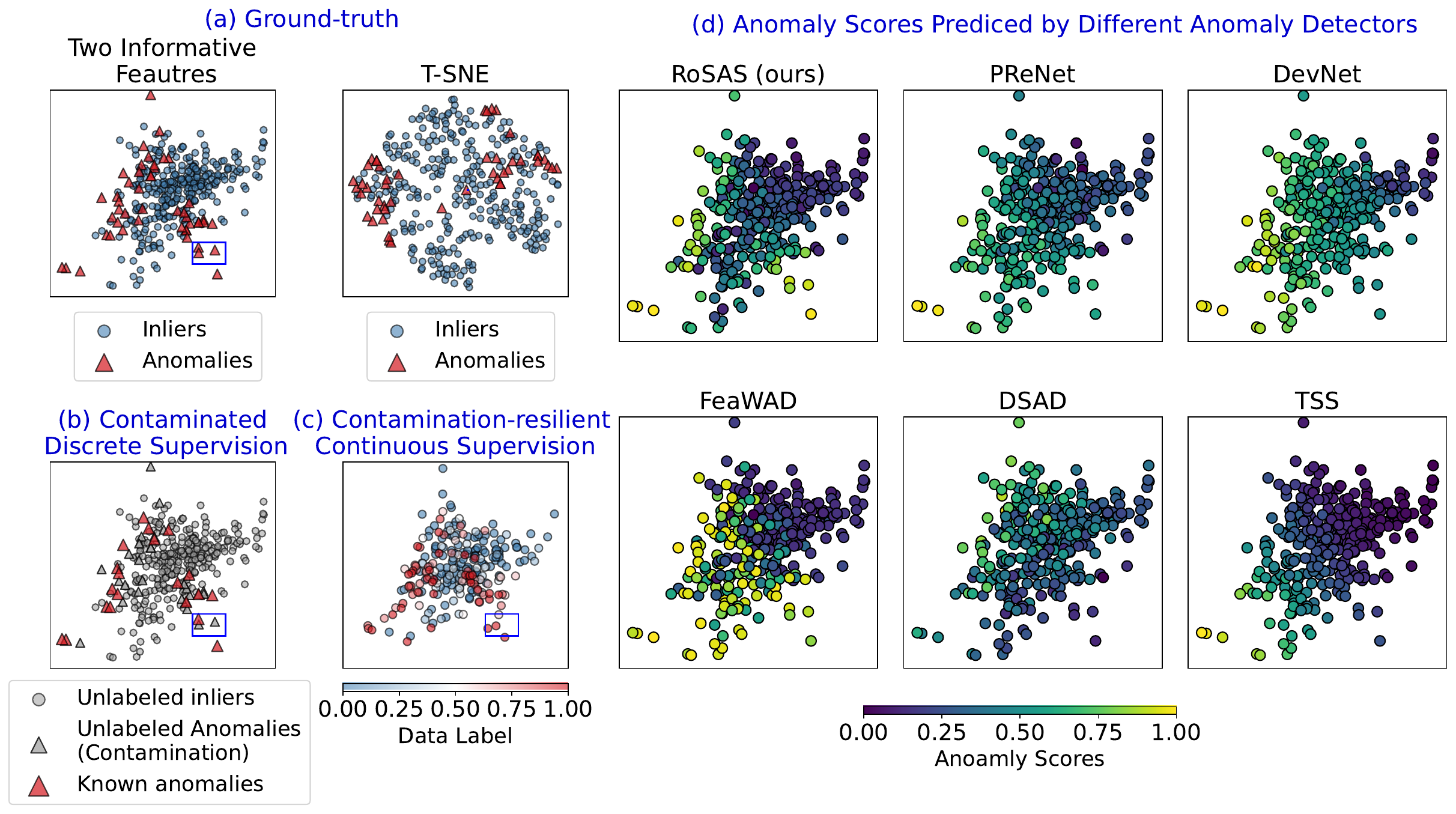}
\caption{ 
	(\textbf{a}) Ground-truth labels of a toy case (the left panel uses two raw features that are informative to show the data distribution, and the right panel shows the 2-D data transformed by T-SNE). In the following sub-figures, we rely on the two raw informative features to visualize. 
	(\textbf{b}) Raw supervision information (i.e., \textit{contaminated discrete supervision}) directly offered by the semi-supervised setting. (\textbf{c})  \textit{Contamination-resilient continuous supervision} generated by our model. (\textbf{d}) Anomaly scoring results of our method RoSAS vs. existing approaches including PReNet \citep{pang2019prenet}, DevNet \citep{pang2019deep,pang2021explainable}, FeaWAD \citep{zhou2021feature}, DSAD \citep{ruff2020deep}, and TSS \citep{zhang2017ccs}. 
	The blue rectangle in (a)(b)(c) is used to highlight two real anomalies that are hidden in the unlabeled set (i.e., anomaly contamination). These two noisy points may mislead the learning model, but they are effectively covered in our supervision. The generated augmented samples in this area are labeled with higher values clearly indicating the anomalism of this field. Benefiting from the proposed contamination-resilient continuous supervision in (c), our method RoSAS produces more accurate anomaly scores than prior arts as shown in (d). 
}
\label{fig:example}
\end{figure}

To exemplify the issues described above, we use a toy dataset\footnote{This toy dataset is generated via the \texttt{make\_classification} function of the \texttt{Scikit-learn} library \citep{scikit-learn}. The dataset is described by ten features, including three informative features, five redundant features (i.e., random linear combinations of the informative features), and two noisy features. The anomaly class contains three clusters.} in Figure \ref{fig:example}. 
Figure \ref{fig:example} (a) visualizes the data with ground-truth annotations, in which the left panel uses the two most relevant dimensions as coordinate axes and the right panel is the T-SNE \citep{van2008visualizing} result.
Most existing models use the contaminated discrete supervisory signals directly supplied by raw labels of the semi-supervised setting, as shown in Figure \ref{fig:example} (b). Data samples in this supervision are labeled by discrete values, and more importantly, this supervision is biased by unlabeled anomalies, i.e., anomaly contamination (e.g., two gray triangles highlighted in the blue rectangle). 
This supervision is not indicative enough to support the detection of the hard anomalies that are mixed up with inliers, or similar to the unlabeled anomalies. As shown in Figure \ref{fig:example} (d), five current state-of-the-art semi-supervised detectors suffer from these issues and fail to yield satisfactory detection results. 

To fill these gaps, this paper further proposes a novel \underline{Ro}bust deep \underline{S}emi-supervised \underline{A}nomaly \underline{S}coring method (termed RoSAS), in which the produced anomaly scores are optimized via \textit{contamination-resilient continuous supervisory signals}.
RoSAS follows our general network structure consisting of a feature representation module and an anomaly scoring module to directly yield anomaly scores, where the whole process is optimized in an end-to-end manner. 
Specifically, we first propose a mass interpolation method to diffuse the abnormality of labeled anomaly examples to the unlabeled area, which yields augmented data samples.
As the interpolation process is measurable according to the diffusion intensity, these newly created data can be labeled with continuous values that faithfully indicate their abnormal degrees, thereby offering continuous supervision to straightly optimize the anomaly scoring mechanism. 
The located area of anomaly contamination can be covered by the new data generated by the interpolation of data combinations with correct labels. 
Even if the anomalies hidden in unlabeled data are used in interpolation, their negative effects can be diluted when they are grouped with genuine normal data or real anomalies in a mass.
Consequently, new supervisory signals can better tolerate the anomaly contamination problem. 
Besides, our optimization process encourages the consistency between the anomaly score of each augmented sample and the score interpolation of their corresponding original data samples. This consistency learning can produce smoother anomaly scores to describe continuous abnormal degrees better. 
Additionally, we pose a feature learning-based objective that ensures effective isolation of labeled anomalies in the intermediate representation, which serves as an optimization constraint to further regularize the network and enhance the robustness w.r.t. anomaly contamination.

Figure \ref{fig:example} (c) illustrates the devised contamination-resilient continuous supervision, which is not only noise-tolerant but very faithful to the ground truth, demonstrating significantly higher supervision quality.
Therefore, as depicted in Figure \ref{fig:example} (d), RoSAS produces more reliable anomaly scoring results than competing methods that rely on raw supervision information.

Our main contributions are summarized as follows.
\begin{itemize}

\item Motivated by the two limitations manifested by the general framework of this research line, we propose a novel semi-supervised anomaly detection method RoSAS, in which we devise a new kind of contamination-resilient continuous supervisory signals to optimize anomaly scores in an end-to-end manner. 

\item We propose a mass interpolation method in RoSAS to generate augmented data samples together with continuous values as data labels. In addition to offering continuous supervision, the created supervisory signals can tolerate anomaly contamination. 

\item We introduce consistency learning which encourages RoSAS to produce smoother anomaly scores, thus better describing abnormal degrees. We also set a feature learning-based objective to regularize RoSAS. The intermediate representation is constrained to further enhance its robustness w.r.t. anomaly contamination.

\end{itemize}

Extensive experiments show that: 1) RoSAS achieves significant AUC-PR and AUC-ROC improvement over state-of-the-art semi-supervised anomaly detection methods; 2) RoSAS obtains more robust and superior performance in settings with different anomaly contamination levels and varying numbers of labeled anomalies. We also empirically show the advantage of the proposed contamination-resilient continuous supervisory signals over discretized, conventional ones and validate contributions of the consistency constraint in anomaly scoring and the regularizer based on feature learning.

\section{Related Work}

This section first reviews unsupervised anomaly detection and summaries semi-supervised models that exploit labeled anomaly examples. 

\subsection{Unsupervised Anomaly Detection}

Traditional unsupervised anomaly detection identifies anomalies according to different data characteristics like proximity and probability \citep{bandaragoda2018isolation,li2020copod,liu2008isolation}.
The burgeoning of deep learning has fueled a plethora of deep anomaly detectors. 
In this research line, many studies \citep{ding2019deep,gong2019memorizing,lv2023adaptive,xu2019mix,zhang2019mscred} train Autoencoders or generative adversarial networks to reconstruct/generate the original inputs. 
Self-supervised methods \citep{golan2018deep,shenkar2022icl,xu2023fascinating} define data-driven supervision and proxy tasks. 
These methods essentially learn intrinsic patterns of training data that are dominated by normal data, and loss values are directly used to estimate abnormal degrees during inference. 
In addition, some studies enhance traditional models by harnessing the strong representation capability of deep learning. Deep SVDD \citep{ruff2018deep} is based on support vector data description \citep{tax2004support}, and DIF \citep{xu2022dif} enhances the isolation process of \citep{liu2008isolation} by proposing deep representation ensemble. 
Basic insights in mainstream deep anomaly detectors can be also achieved via non-deep models. 
The literature \citep{xu2021reconstruction} uses tree models to realize the reconstruction pipeline. 
Although these unsupervised methods are intuitive and practical, without knowing real anomalies, they often lead to many false alarms which may overwhelm anomalies of real interest.

\subsection{Semi-supervised Anomaly Detection}

In contrast, relatively few studies consider semi-supervised anomaly detection utilizing limited anomaly examples. In this category, we also review the related literature that uses both labeled normal data and labeled anomalies since they can also work under this scenario by treating unlabeled data as normal.

The study \citep{zhang2018anomaly} employs canonical clustering to divide labeled anomalies into $k$ clusters and detect anomalies by a ($k$+1)-class classifier.
Non-deep unsupervised anomaly detection methods can be also enhanced to leverage weak incomplete supervision. \cite{barbariol2022tiws} extend ensemble-based isolation forest \citep{liu2008isolation} by leveraging supervision information to filter ensemble members, which improves detection performance and simultaneously reduces computational costs.

This incomplete supervision can be also leveraged in deep models to learn a good representation. 
Some methods map input data to a representation space and explicitly impose specific criteria such as triplet loss \citep{pang2018learning} and anomaly-informed one-class loss \citep{ruff2020deep} upon the representation. They further employ distance-based anomaly scoring protocols upon this learned representation space.
Besides, data representations can be also implicitly learned via Autoencoders or generative adversarial networks. 
\cite{huang2020esad} propose a novel encoder-decoder-encoder structure. It modifies the reconstruction loss to force the network to reconstruct labeled anomalies to pre-defined noises. 
Bidirectional GAN is used in \citep{tian2022bigan}, in which labeled anomalies are used to learn a probability distribution, and the distribution can assign low-density values to labeled anomalies. 
These methods are indirectly optimized to yield abnormal degrees of data samples, and anomaly scores can be only obtained in an isolated manner.

Some advanced deep approaches are in an end-to-end fashion to directly optimize the produced anomaly scores.
The pioneering work in this research line \citep{pang2019deep,pang2021explainable} assumes anomaly scores of normal data follow a Gaussian distribution and yield the reference score. 
It further employs the z-score function to define the deviation loss to ensure anomaly scores of labeled anomalies significantly deviate from the reference. 
An Autoencoder is added to the above framework in \citep{zhou2021feature}. In addition to a deviation loss imposed on the derived anomaly scores, the reconstruction error of labeled anomalies is optimized to be as larger as a pre-defined margin. 
By defining the ordinal target of paired data samples, \cite{pang2019deep} use mean absolute error to optimize anomaly scores. The cross-entropy loss is used in \citep{ding2022catching} to classify labeled anomalies, transferred pseudo anomalies, and latent residual anomalies from unlabeled data.

It is also noteworthy that, except for tabular data or images, related studies also consider this semi-supervised learning paradigm of anomaly detection in graph data \citep{ding2021few,dou2020enhancing,zhou2022unseen} and time series \citep{carmona2022neural,huang2022semi}.

\section{A General Framework of Deep Semi-supervised \\Anomaly Detection}\label{sec:framework}

\paragraph{Problem Statement}
We assume a few labeled anomaly examples $\mathcal{X}_A$ are accessible in addition to large-scale unlabeled training data $\mathcal{X}_U$, where $|\mathcal{X}_A| \ll |\mathcal{X}_U| $, i.e., the quantity of labeled anomalies is very small compared to the number of true anomalies and the whole dataset. 
Given the training data $\mathcal{X} = \mathcal{X}_U \cup \mathcal{X}_A$, an anomaly detection model is trained to assign higher scores to data samples with higher likelihoods to be anomalies.

\subsection{General Framework}

We below introduce a general framework of deep semi-supervised anomaly detection, and this framework can well cover representative existing models \citep{carmona2022neural,pang2018learning,pang2019deep,pang2021explainable,pang2019prenet,ruff2020deep,wu2021surrogate,zhou2021feature} and summarize their limitations.

We first define the network structure of the framework. Let $f: \mathcal{X} \mapsto \mathbb{R}$ represent the network that outputs anomaly scores given the input data $\mathcal{X}$. The whole procedure can be divided into a feature representation module $\phi: \mathcal{X} \mapsto \mathbb{R}^{H}$ and an anomaly scoring module $\psi:\mathbb{R}^{H} \mapsto \mathbb{R}$. Feature representation module $\phi$ aims to map $\mathcal{X}$ into a feature space with dimensionality $H$. Anomaly scoring module $\psi$ outputs final anomaly scores based on the intermediate representation. Anomaly detection network $f$ is denoted as:
\begin{equation}\label{eqn:e2e}
f(\mathbf{x}) = \psi \big( \phi ( \mathbf{x} ; \Theta_\phi \big) ; \Theta_\psi ),
\end{equation}
where $\Theta_\phi$ and $\Theta_\psi$ are network parameters in $\phi$ and $\psi$.

We then define a general learning objective. 
Under the semi-supervised setting, each data sample in the training set can be assigned a target. 
Let $\mathcal{D} = \{(\mathbf{x}, y) \in \mathcal{X} \times \mathcal{Y} \}$ with $\mathcal{Y}=\{y^+, y^-\}$ be a set of training samples, where $y^+$ denotes labeled anomalies and $y^-$ denotes unlabeled data. Although most of $(\mathbf{x}, y^-)$ are genuine normal samples, there are still some unlabeled anomalies that are wrongly assigned $y^-$. Data augmentation techniques can be also used to obtain a new training set $\tilde{\mathcal{D}} = \{ (\tilde{\mathbf{x}}, \tilde{y})\}$.
A general objective function is defined as follows:
\begin{equation}
\small
\begin{aligned}
	\min_{ \{\Theta_\phi, \Theta_\psi \} } & \text{  }
	\mathbb{E}_{(\mathbf{x} , y) \sim \mathcal{D}  } 
	\Big[ \mathcal{L}_D \big( \psi(\phi(\mathbf{x})), y  \big)  \Big] 
	+
	\mathbb{E}_{(\mathbf{x} , y) \sim \mathcal{D} } 
	\Big [\mathcal{L}'_D \big( \phi(\mathbf{x}), y \big) \Big]     \\
	+ & \mathbb{E}_{(\tilde{\mathbf{x}}, \tilde{y}) \sim \tilde{\mathcal{D}} }
	\Big [\mathcal{L}_{\tilde{D}} \big (\psi(\phi(\tilde{\mathbf{x}})), \tilde{y} \big) \Big] +
	\mathbb{E}_{(\tilde{\mathbf{x}}, \tilde{y}) \sim \tilde{\mathcal{D}} }
	\Big [\mathcal{L}'_{\tilde{D}} \big (\phi(\tilde{\mathbf{x}}), \tilde{y} \big) \Big]  .
\end{aligned}\label{eqn:obj}
\end{equation}
The above equation can be interpreted as the optimization of the representation $\phi(\cdot)$ and/or the final anomaly scores $\psi(\phi(\cdot))$ by using supervision signals provided by original data $\mathcal{D}$ and/or augmented data $\tilde{\mathcal{D}}$.

\subsection{Generalization of Current Studies}
As for the network structure in Eqn (\ref{eqn:e2e}), different network structures are used according to data types and data characteristics, e.g., multi-layer perceptron net is used for multi-dimensional tabular data \citep{pang2018learning,pang2019deep,pang2021explainable,pang2019prenet,wu2021surrogate,zhou2021feature}, convolutional net is used for image data \citep{ruff2020deep}, and temporal net is used for time series \citep{carmona2022neural}.

The proposed objective function Eqn. (\ref{eqn:obj}) can well cover existing deep semi-supervised anomaly detectors by specifying each of its terms, as shown in Table \ref{tab:framework}. We below explain their instantiation method in detail.
\begin{itemize}
\item Deep SAD \citep{ruff2020deep} defines $\mathcal{L}'_D$. 
Upon the representation space, labeled anomalies are repulsed to be distant to a pre-defined center $\mathbf{c}$ as far as possible, and unlabeled data are expected to be included in a compact hypersphere with the minimum volume taking $\mathbf{c}$ as the center.
\item FeaWAD \citep{zhou2021feature} first instantiates $\mathcal{L}_D$. It is optimized to enlarge anomaly scores of labeled anomalies to a pre-defined margin $e$ and maps scores of unlabeled data to zero. $\mathcal{L}'_D$ is further instantiated by a reconstruction loss with the help of an Autoencoder structure. 
\item DevNet \citep{pang2019deep,pang2021explainable} specifies $\mathcal{L}_D$. It proposes a z-score-based deviation function by assuming a pre-defined Gaussian prior of anomaly scores and sampling reference scores $\mu$ and standard deviation values $\sigma$ from this distribution.  
\item PReNet \citep{pang2019prenet} specifies $\mathcal{L}_{\tilde{D}}$ as Mean Absolute Error (MAE) between the scores of concatenated pairs (anomaly-unlabeled, anomaly-anomaly, and unlabeled-unlabeled) and pre-defined ordinal regression targets ($e_1$, $e_2$, and $e_3$).
\end{itemize}
REPEN \citep{pang2018learning} and NCAD \citep{carmona2022neural} also repulse labeled anomalies upon the representation space, as has been done in Deep SAD. PLSD \citep{wu2021surrogate} is similar to PReNet by replacing MAE loss with cross-entropy loss. Therefore, these methods are omitted in Table \ref{tab:framework}.

\begin{table*}[htbp]
\centering
\caption{Instantiation method and gaps of existing deep semi-supervised anomaly detection studies}
\scalebox{0.62}{
	\begin{tabular}{p{4.3cm}  p{4.7cm}  p{4.7cm} p{2.1cm}<{\centering} p{2.2cm}<{\centering}}
		\toprule
		
		\textbf{\makecell[l]{Anomaly \\ Detectors}} & 
		\textbf{\makecell[c]{$\mathcal{L}_D$ / $\mathcal{L}_{\tilde{D}}$: \\anomaly score \\ optimization}} &
		\textbf{\makecell[c]{$\mathcal{L}'_D$ / $\mathcal{L}'_{\tilde{D}}$: \\ representation \\ optimization}}   &
		\textbf{Robustness} &  
		\textbf{\makecell[c]{Continuous \\ supervision}} 
		\\
		\midrule
		
		%
		
		\makecell[l]{Deep SAD \\ \citep{ruff2020deep}} &  
		\multicolumn{1}{c}{-} &
		\makecell[l]{
			$\mathbbm{1}_{y^+} 
			\Vert \phi(\mathbf{x}) \!-\! \mathbf{c} \Vert^{-1} +$\\
			$\mathbbm{1}_{y^-}\Vert \phi(\mathbf{x}) \!-\! \mathbf{c} \Vert$ }
		& \XSolidBrush  &  \XSolidBrush  \\
		
		\specialrule{0em}{4pt}{2pt}
		\makecell[l]{FeaWAD \\ \citep{zhou2021feature}} &
		\makecell[l]{ 
			$\mathbbm{1}_{y^+}\max(0, e \!-\! \psi(\phi(x))) +$\\ $\mathbbm{1}_{y^-}  \vert \psi(\phi(\mathbf{x})) \vert $}
		&
		\makecell[l]{
			$\mathbbm{1}_{y^+} \max(0, e \!-\! \Vert \phi(\mathbf{x})  \!-\! \mathbf{x} \Vert ) + $ \\
			$\mathbbm{1}_{y^-}  \Vert \phi(\mathbf{x}) \!-\! \mathbf{x} \Vert $ }
		& \XSolidBrush &  \XSolidBrush
		\\
		
		\specialrule{0em}{4pt}{2pt}
		\makecell[l]{DevNet \\ \citep{pang2019deep,pang2021explainable}} &  	
		\makecell[l]{
			$ \mathbbm{1}_{y^+}
			\max(0, e \!-\! \frac{\psi(\phi(\mathbf{x})) - \mu}{\sigma})+$\\  $
			\mathbbm{1}_{y^-} 
			\vert \frac{\psi(\phi(\mathbf{x})) - \mu}{\sigma} \vert $}  
		& \multicolumn{1}{c}{-}
		& \Checkmark &  \XSolidBrush
		\\ 
		
		\specialrule{0em}{5pt}{2pt}
		\makecell[l]{PReNet \\ \citep{pang2019prenet}} &  
		\makecell[l]{
			$ \mathbbm{1}_{\langle y_i^+ \!, y_j^+ \rangle}\vert \psi(\phi( \langle \mathbf{x}_i, \!\mathbf{x}_j \rangle)) \!-\! e_1 \vert  + $ \\ 
			$ \mathbbm{1}_{\langle  y_i^+ \!,  y_j^- \rangle}\vert 	\psi(\phi( \langle \mathbf{x}_i,\! \mathbf{x}_j \rangle)) \!-\! e_2 \vert + $ \\ 
			$
			\mathbbm{1}_{\langle y_i^- \!, y_j^- \rangle}\vert 	\psi(\phi( \langle \mathbf{x}_i,\! \mathbf{x}_j \rangle)) \!-\! e_3 \vert
			$ } & \multicolumn{1}{c}{-}
		& \Checkmark &  \XSolidBrush
		\\
		
		
		\midrule
		
		\textbf{RoSAS (ours)} & 
		
		\makecell[l]{$\ell \big( \psi(\phi(\tilde{\mathbf{x}})), \tilde{y} \big)
			+ $ \\ $
			\ell \big( \psi (\phi ({\tilde{\mathbf{x}}})), \sum_{i=1}^{k} \lambda_i \psi(\phi(\mathbf{x}_i))  \big)$ }
		&
		\makecell[l]{$\max(0, e+d(\phi(\mathbf{x}^-), \phi(\mathbf{q}))) - $ \\ $d(\phi(\mathbf{x}^+), \phi(\mathbf{q})) $} 
		& \Checkmark & \Checkmark
		\\

		\bottomrule
		
	\end{tabular}%
}
\label{tab:framework}%
\end{table*}%

\subsection{Limitations of Current Studies}\label{sec:limitation}
By looking into Table \ref{tab:framework}, we perceive two key gaps in these existing approaches, i.e., \textit{robustness w.r.t. contamination} and \textit{continuous supervision of optimization}.

\subsubsection{Robustness w.r.t. contamination}
Deep SAD and FeaWAD use unlabeled data as an opposite data class against labeled anomalies. They define a specific loss term (starting with $\mathbbm{1}_{y^-}$) to \textit{indistinguishably} map all of these unlabeled data to a unified target. 
This operation seems to be reasonable due to the unsupervised nature of anomaly detection (i.e., anomalies are rare data). 
To further enhance detection performance, we need to consider the negative effect brought by anomaly contamination in unlabeled data and improve the model robustness. 
DevNet \citep{pang2019deep,pang2021explainable} assumes a Gaussian distribution prior of optimization targets. Due to the flexibility of Gaussian distribution, it can partially eliminate interference.
PReNet \citep{pang2019prenet} uses vector concatenation of data pairs to redefine three surrogate classes. This kind of data combination can resist small anomaly contamination since the interference of noisy samples can be mitigated when they are combined with genuine normal data or labeled anomalies. 

\subsubsection{Continuous supervision}
Anomaly scores produced by anomaly detection models are expected to indicate abnormal degrees, and human investigators can examine the reported suspicious data in descending order of anomaly scores.
Deep SAD, DevNet and FeaWAD utilize discrete binary supervision to respectively map labeled anomalies and unlabeled data to \textit{two extremes} during training, but their models are required to output continuous anomaly scores during inference. 
Specifically, Deep SAD uses one fixed center $\mathbf{c}$ (unlabeled data are gathered at this center, and anomalies are repelled), FeaWAD directly maps anomaly scores to zero and a pre-defined margin $e$, and DevNet first calculates z-scores of anomaly scores and employs zero and a margin $e$ as two extreme targets. 
Instead of using two extremes, PReNet employs three pre-defined ordinal targets, but this is also a kind of discrete supervision. 
Prior arts utilize the above discrete supervision information to optimize the continuously distributed anomaly scores.
Due to the lack of continuous supervision, these models may fail to learn how to subtly describe abnormal degrees, resulting in suboptimal anomaly scoring mechanism.





\section{The proposed RoSAS}

\begin{figure}[t]
\centering	\includegraphics[width=0.8\textwidth]{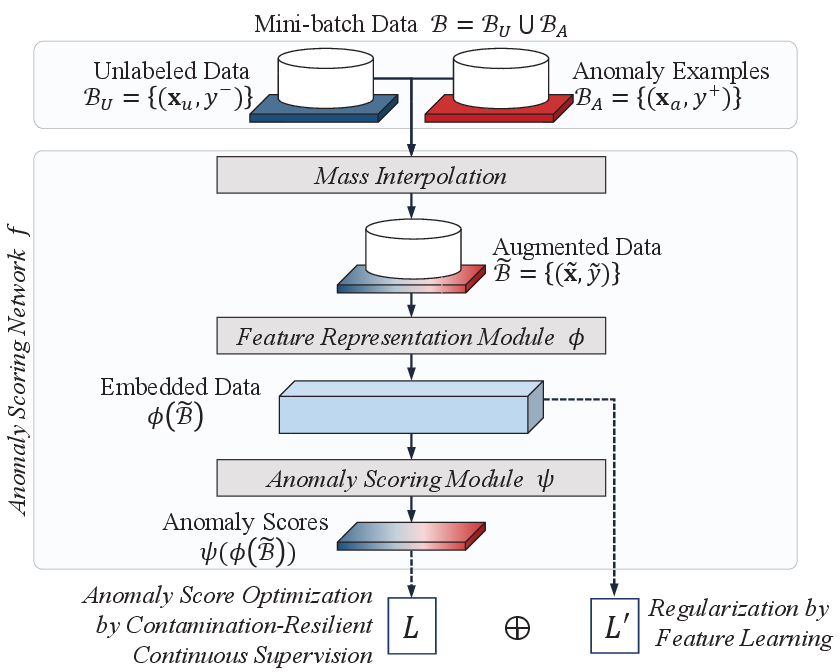}
\caption{The overall procedure of RoSAS. Each training mini-batch $\mathcal{B}$ is composed of unlabeled data $\mathcal{B}_U$ and anomaly examples $\mathcal{B}_A$. 
	The derived anomaly scores are end-to-end optimized by loss function ${L}$. ${L}$ is defined based on contamination-resilient continuous supervision signals that are offered by augmented samples $\tilde{\mathcal{B}}$. A feature learning-based objective $L'$ on the intermediate representation $\phi$ is added to further regularizes the network. ${L}$ and ${L}'$ are assembled via dynamic weight averaging $\oplus$. 
}
\label{fig:rosas}
\end{figure}

This paper proposes a concrete deep semi-supervised anomaly detection method termed RoSAS.
The overall procedure is shown in Figure \ref{fig:rosas}. As described in Eqn. (\ref{eqn:e2e}), RoSAS also follows the basic network structure $f$ consisting of a feature representation module $\phi(\cdot|\Theta_\phi)$ and a scoring module $\psi(\cdot|\Theta_\psi)$. RoSAS is optimized by the loss function ${L}$ and the regularizer ${L}'$.

The network architecture of $\phi$ and $\psi$ is determined according to the input data types and/or data characteristics.
In terms of the design of loss function $L$, the simplest way is to directly treat the whole unlabeled set as normal data, and discrete targets can be assigned to labeled anomaly examples and unlabeled data, as has been done in many prior studies \citep{pang2019deep,zhou2021feature}. 
However, these labels are inaccurate due to anomaly contamination and fail to sufficiently reflect anomaly scores that by definition take on a continuous distribution.
It is very challenging to obtain reliable abnormal degrees of original training data since we do not exactly know whether one anomaly is more abnormal than another. 
Hence, we finally resort to synthesizing new data samples by diffusing the abnormality of these accessible labeled anomalies to the unlabeled area, and thus the abnormal degree is controllable. 
That is, the design of $L$ is essentially to specify the term $\mathcal{L}_{\tilde{D}}$ in Eqn. (\ref{eqn:obj}). Specifically, based on the original training mini-batch data $\mathcal{B}$, we propose a mass interpolation method to create a set of augmented data samples attached with continuous supervision targets $\tilde{\mathcal{B}}=\{ ( \tilde{\mathbf{x}}, \tilde{y} ) \}$.
Note that this new supervision can also resist the anomaly contamination problem. 
The contaminated area can be covered by new data samples generated via combinations of data with correct labels. 
Anomaly contamination also becomes less harmful when these notorious unlabeled anomalies are combined with genuine normal samples or labeled anomalies during the interpolation. 
Consequently, RoSAS successfully devices contamination-resilient, continuous supervision of anomaly score optimization.

Further, motivated by the potential generalization and regularization effect of multi-task learning \citep{vandenhende2021multi},
we define an additional objective ${L}'$ upon the feature representation module $\phi$ to encourage significant deviations of labeled anomalies in the intermediate representation space. The network can be regularized by this new optimization constraint, which further improves the robustness to anomaly contamination. 

Two loss terms $L$ and $L'$ are finally assembled via dynamic weight averaging $\oplus$ to avoid manually setting a fixed weight. Dynamic weight averaging $\oplus$ can balance the optimization pace at two loss terms.

\vspace{+2mm}

We below present the design of the loss function $L$ (Section \ref{sec:loss}), the regularization term $L'$ (Section \ref{sec:regularization}), and the dynamic averaging $\oplus$ (Section \ref{sec:dynamic}) in detail. We finally illustrate the procedure of RoSAS by giving its pseudo code (Section \ref{sec:algo}).

\subsection{Anomaly Score Optimization by Contamination-resilient\\ Continuous Supervision} \label{sec:loss}

RoSAS first produces new augmented data samples with controllable and reliable abnormal degrees via the mass interpolation method.
Compared to directly using contaminated discrete targets, RoSAS can optimize anomaly scores as a regression problem with faithful continuous targets.

Specifically, based on the original mini-batch data $\mathcal{B}=\{(\mathbf{x}, y)\}$ with $y=1$ for labeled anomalies and $y=-1$ for unlabeled data, RoSAS creates a novel mini-batch $\tilde{\mathcal{B}}$ of augmented data samples by the mass interpolation.
These augmented data samples are synthesized as a weighted summation of $k$ original data samples.
Different weights of candidates $\{\lambda_1, \cdots, \lambda_k\}$ produce continuous targets in new supervision. $\tilde{\mathcal{B}}$ is defined as follows.
\begin{equation}\label{eqn:tildeb}
\begin{split}
	\small
	\tilde{\mathcal{B}} = \Big\{ (\tilde{\mathbf{x}}, \tilde{y} ) | 
	\tilde{\mathbf{x}} = 
	\textstyle\sum_{i=1}^{k} \lambda_i \mathbf{x}_i, 
	\tilde{y} = \textstyle\sum_{i=1}^{k} \lambda_i y_i \Big \}
	,
\end{split}
\end{equation}
where $\sum_{i=1}^{k}\lambda_i = 1$, $\{(\mathbf{x}_i, y_i)\}_{i=1}^{k} \subset \mathcal{B} $, and $\lambda_i$ is sampled from a continuous distribution.

As for the distribution of $\lambda$, inspired by \citep{zhang2018mixup}, RoSAS uses Beta distribution, i.e., $\lambda \sim Beta(\alpha, \alpha)$. It is because adjusting distribution parameter $\alpha$ can produce different types of weights, e.g., a uniform distribution when $\alpha=1$ or an approximate truncated normal distribution when $\alpha$ is a larger value. 
If $\alpha > 1$, interpolation weights will centralize around 0.5, and some noisy labels might be produced (e.g., mixing two anomalies may yield a new sample in the normal manifold, but an anomalous label is given to this new sample). This is also known as ``manifold intrusion'' \citep{guo2019mixup}. To tackle this problem, we use $\alpha = 0.5$ by default. In doing so, interpolation weights are more likely to be slightly larger/smaller than 0/1, which makes the interpolation located in the local regions of the original samples. Thus, these possible noisy labels can be reduced or eliminated.

The loss function $L$ measures the empirical risks of derived anomaly scores of augmented samples compared to the continuous targets. Additionally, we add a consistency term to measure the difference between each augmented sample's anomaly score and the weighted summation of their original data instances' anomaly scores using the same interpolation weights. This consistency learning is to encourage the network to produce smoother anomaly scores, thus better describing abnormal degrees. Therefore, $L$ is finally defined as:
\begin{equation}\label{eqn:lphi}
\begin{split}
	\small
	L =
	\mathbb{E}_{( \tilde{\mathbf{x}}, \tilde{y} ) \sim \tilde{\mathcal{B}}}
	\Bigg[ \ell \bigg( \psi(\phi(\tilde{\mathbf{x}})), \tilde{y} \bigg)
	+ 
	\ell \bigg( \psi (\phi ({\tilde{\mathbf{x}}})), \sum_{i=1}^{k} \lambda_i \psi(\phi(\mathbf{x}_i))  \bigg)
	\Bigg]
	,
\end{split}
\end{equation}
where $\{\mathbf{x}_i\}_{i=1}^{k}$ is the original data samples when creating $\tilde{\mathbf{x}}$ as defined in Eqn. (\ref{eqn:tildeb}), and $\ell(\cdot, \cdot)$ is a base regression loss.


The loss function $L$ not only fulfills continuous optimization but can tolerate anomaly contamination. 
The unlabeled set is still dominated by genuine normal data because of the rarity of anomalies. 
The contaminated area can be calibrated via new data samples that are augmented from a group of data with correct labels.
Even if these noisy unlabeled anomalies are sampled in Eqn. (\ref{eqn:tildeb}), they are likely to be combined with labeled anomalies or real normal data. That is, the generation process of augmented data samples also dilutes the anomaly contamination in a simple yet effective manner. 
Therefore, RoSAS is more robust w.r.t. anomaly contamination.

\subsection{Regularization by Feature Learning}\label{sec:regularization}
The feature representation module $\phi: \mathcal{X} \mapsto \mathbb{R}^H$ maps input data into a new feature space. We further define a new loss term ${L}'$ upon this intermediate representation space, which serves as a new optimization constraint to regularize the network and further enhance the robustness.

To fully leverage these labeled anomalies, $L'$ is designed to learn a feature representation that can effectively repulse these labeled anomaly examples from unlabeled data (the majority of unlabeled data is normal).
Let $\mathbf{q}$ be an anchor data object, and we utilize the difference between the deviation of unlabeled-anchor and anomaly-anchor pairs to measure the separability of labeled anomalies, which is defined as follows. 
\begin{equation}\label{eqn:lf}
\small
{L}'\! =\!  \mathbb{E}_{
	\substack {(\mathbf{x}^{+}, y^{+}) \sim \mathcal{B}_A \\
		(\mathbf{x}^{-}, y^{-}) \sim \mathcal{B}_U}
}\!
\bigg [
\!\max\! \Big( d \big( \phi(\mathbf{x}^{-}) , \phi(\mathbf{q})\big)  - d \big( \phi(\mathbf{x}^{+}) , \phi(\mathbf{q}) \big) \!+\! e , 0 \Big)  \bigg] ,
\end{equation}
where $d(\cdot , \phi(\mathbf{q}))$ indicates the deviation given the anchor data, and $e$ is a margin. 
Different distance functions or similarity measures can be used. Considering the simplicity, we employ Euclidean distance here. 
$\mathcal{B}_{U}$ and $\mathcal{B}_{A}$ are unlabeled data and labeled anomalies in mini-batch $\mathcal{B}$. 
In practical implementation, a mini-batch of anchor data is sampled from the unlabeled set along with mini-batch $\mathcal{B}$, i.e., $\mathbf{q} \in \mathcal{B}_q, \mathcal{B}_q \subset\mathcal{X}_u$. Anchor data can also be determined as representative normal prototypes if labeled normal data are available.

It is noteworthy that ${L}'$ uses a relative and soft manner to judge whether these labeled anomalies are effectively separated by introducing a reference divergence degree $d(\phi(\mathbf{x}^{-}) , \phi(\mathbf{q}))$ between unlabeled data and anchor data. It avoids blindly enlarging $d(\phi(\mathbf{x}^{+}) , \phi(\mathbf{q}))$, i.e., the anomalies that have been successfully deviated are no longer required to be optimized; thus, the optimizer can focus on true errors. 
On the other hand, even if unlabeled anomalies are wrongly identified as anchor data $\mathbf{q}$ or $\mathbf{x}^-$ in Eqn. \ref{eqn:lf}, this function can still work to isolate labeled anomalies. 

\subsection{Dynamic Averaging}\label{sec:dynamic}
Instead of setting a fixed weight, the loss term ${L}$ and the regularizer ${L}'$ are assembled via dynamic weight averaging \citep{liu2019end}, i.e., 
\begin{equation}
wL + (1-w)L',
\end{equation}
where $w$ is defined according to the optimization pace (loss descending rate) of ${L}$ and ${L}'$. 
$w$ is defined as follows. 
\begin{equation}\label{eqn:weight}
w = \frac{\exp( {L}/T \bar{L})}
{ \exp( L/T\bar{L}) + \exp( L'/T\bar{L}')   },
\end{equation}
where $\bar{L}$ and $\bar{L}'$ are the average loss of the last training epoch, and $T$ is the temperature as used in the softmax function.

\subsection{Algorithm of RoSAS}\label{sec:algo}

Algorithm \ref{alg:RoSAS} presents the training procedure of RoSAS. Step 1 initializes the loss terms for the subsequent dynamic weight averaging. For each training batch, a mini-batch of known anomalies $\mathcal{B}_a$ of size $b$ is sampled from $\mathcal{X}_A$, $2b$ data objects are sampled from $\mathcal{X}_U$ as mini-batch $\mathcal{B}_u$ and anchor $\mathcal{B}_q$ in Steps 4-5. Step 6 creates a mini-batch of augmented data. The scoring loss and the regularization term are computed in Steps 7-8. Dynamic weights are adjusted in Step 9. Step 10 performs back propagation to optimize the network parameters w.r.t. the loss $wL + (1-w) L'$. Step 12 updates the average losses.

The computation of loss term ${L}$ and ${L}'$ has an overall time complexity of $O(n\_epoch * n\_batch * b * H)$, where $H$ is the representation dimension. The time complexity of RoSAS also depends on the network structure. We take a multi-layer perceptron network with $u$-hidden layer as an example, the feed-forward propagation incurs $O(n\_epoch * n\_batch * b * (Dh_1 + h_1h_2+\dots+h_u*1))$, where $h_i$ is the number of hidden units in the $i$-th hidden layer.

\renewcommand{\algorithmicrequire}{\textbf{Input:}}
\renewcommand{\algorithmicensure}{\textbf{Output:}} 
\begin{algorithm}[htbp]
\caption{Training of RoSAS}
\label{alg:RoSAS}
\begin{algorithmic}[1] 
	\REQUIRE  Labeled anomaly examples - $\mathcal{X}_A$, unlabeled data - $\mathcal{X}_U$
	\ENSURE Anomaly Scoring network -  $\psi(\phi(\cdot ; \Theta_\phi); \Theta_\psi )$
	\STATE Initialize $\bar{{L}} \leftarrow 1$, $\bar{L}' \leftarrow 1$
	\FOR{$t=1$ to $n\_epoch$}
	\FOR{$j=1$ to $n\_batch$}
	\STATE $\mathcal{B}_A \!\leftarrow$ randomly sample $b$ data objects from $\mathcal{X}_A$ 
	\STATE $\mathcal{B}_U, \mathcal{B}_q \! \leftarrow$ randomly sample $2b$ data objects from $\mathcal{X}_U$ 
	\STATE $\tilde{\mathcal{B}}\leftarrow$ create augmented mini-batch by Eqn. (\ref{eqn:tildeb})
	\STATE Compute loss ${L}$ by Eqn. (\ref{eqn:lphi})
	\STATE Compute regularizer ${L}'$ by Eqn. (\ref{eqn:lf})
	\STATE Compute weights $w$ by Eqn. (\ref{eqn:weight}) 
	\STATE Optimize parameters $\{\Theta_\phi, \Theta_\psi\}$ w.r.t. $w {L} + (1-w){L}'$
	\ENDFOR 
	\STATE $\bar{L}, \bar{L'} \leftarrow$ average loss over the current epoch 
	\ENDFOR
	\STATE \textbf{return} $\psi(\phi(\cdot))$
\end{algorithmic}
\end{algorithm}

\section{Experiments}
In this section, we first describe experimental setup (Section \ref{sec:expsetup}) and conduct experiments to answer the following questions:


\begin{itemize}
\item \textbf{Effectiveness}: Is RoSAS more effective than state-of-the-art anomaly detectors on real-world datasets? Can RoSAS handle different types of anomalies? (Section \ref{sec:effectiveness})

\item \textbf{Robustness}: How does the robustness of RoSAS and its competitors when the unlabeled set is contaminated by different levels of anomalies? (Section \ref{sec:robustness})

\item \textbf{Data Efficacy}: Can RoSAS fully leverage different numbers of labeled anomalies? (Section \ref{sec:efficacy})

\item \textbf{Scalability Test}: How does the time efficiency of RoSAS compared to its competitors? (Section \ref{sec:scalability})

\item \textbf{Ablation Study}: Do key designs contribute to better anomaly detection performance? (Section \ref{sec:ablation})

\item \textbf{Sensitivity}: How do the hyper-parameters influence the detection performance of RoSAS? (Section \ref{sec:sensitivity})

\end{itemize}

\subsection{Experimental Setup}\label{sec:expsetup}

\subsubsection{Datasets}
Eleven publicly available real-world datasets are used\footnote{These datasets are available at \url{https://github.com/GuansongPang/ADRepository-Anomaly-detection-datasets}, \url{https://www.unb.ca/cic/datasets/}, and \url{http://odds.cs.stonybrook.edu/}}. The dataset information is reported in Table \ref{tab:dataset}, including abbreviation (Abbr.), domain/task, data dimensionality ($D$), the number of data samples ($N$), and the anomaly ratio ($\delta$).
The first eight datasets are with real anomalies, which cover three important real-world applications of anomaly detection in cybersecurity, medicine, and finance. 
The last three datasets are from ODDS, a popular repository of anomaly detection datasets, and they contain semantic anomalies. All of these datasets are broadly used as benchmarks in many anomaly detection studies, e.g., \citep{bandaragoda2018isolation,pang2019deep,xu2022dif}. 
We scale each feature to $[0,1]$ via min-max normalization. All the datasets are separated by a random 60:20:20 train-valid-test split while maintaining the original anomaly proportion.

\begin{table}[t]
\centering
\caption{Dataset information. Abbr. is the dataset abbreviation used in the following experiments. $D$ and $N$ denote data dimensionality and data size per dataset, respectively. $\delta$ indicates the anomaly ratio. }
\scalebox{0.86}{
	\begin{tabular}{llllll}
		\toprule
		\textbf{Data} & \textbf{Abbr.} & \multicolumn{1}{l}{\textbf{Domain/Task}} & 
		\multicolumn{1}{l}{\textbf{$D$}} &
		\multicolumn{1}{l}{\textbf{$N$}} &
		\textbf{ $\delta$ }
		\\
		\midrule
		
		CIC-DoHBrW2020 & DoH   & Intrusion Detection & 29    & 1,167,136 & 21.4\% \\
		CIC-IDS2017 WebAttack & WebAttack & Intrusion Detection & 78    & 700,284 & 0.3\% \\
		CIC-IDS2017 PortScan & PortScan & Intrusion Detection & 78    & 816,385 & 19.5\% \\
		UNSW-NB15 Exploit & Exploit & Intrusion Detection & 196   & 96,000 & 3.1\% \\
		UNSW-NB15 Backdoor & Backdoor & Intrusion Detection & 196   & 95,329 & 2.4\% \\
		Thyroid disease & Thyroid & Disease Diagnosis & 21    & 7,200 & 7.4\% \\
		KDD Cup 2014 Donors & Donars & Funding Prediction & 10    & 619,326 & 5.9\% \\
		Credit card fraud detection & Fraud & Fraud Detection & 29    & 284,807 & 0.2\% \\
		Covertype & Cover & Ecosystem & 10    & 286,048 & 1.0\% \\
		Letter recognition & Letter & Recognition & 32    & 1,600 & 6.3\% \\
		Pen-based recognition & Pendigits & Recognition & 16    & 6,870 & 2.3\% \\

		\bottomrule
	\end{tabular}%
}
\label{tab:dataset}%
\end{table}%

\subsubsection{Competitors}

We employ ten anomaly detection models from three categories as competing methods of RoSAS:
\begin{itemize}
\item \textit{Semi-supervised Anomaly Detector}: Five deep semi-supervised anomaly detection methods including PReNet \citep{pang2019prenet}, FeaWAD \citep{zhou2021feature}, DevNet \citep{pang2019deep,pang2021explainable}, Deep SAD (DSAD for short) \citep{ruff2020deep}, and BiGAN \citep{tian2022bigan} are used.
TiWS-iForest (WSIF for short) \citep{barbariol2022tiws} is an enhanced version of \citep{liu2008isolation}, which leverages weak supervision to improve detection performance. 
These competitors fall into different categories of existing techniques, representing the state-of-the-art performance of this semi-supervised setting. 

\item \textit{PU learning-based Method}: Learning from positive and unlabeled data (PU learning) is also a related field if we treat anomalies as positive data. We choose a representative PU learning-based anomaly detector \citep{zhang2017ccs} as our competitor, which combines the two-stage strategy and the cost-sensitive strategy (TSS for short).

\item \textit{Unsupervised Anomaly Detector}: DIF \citep{xu2022dif}, IF \citep{liu2008isolation}, and COP \citep{li2020copod} are employed.
DIF is an isolation-based method that is empowered by deep representation ensemble. 
IF is a popular anomaly detection algorithm that is broadly used in many industrial applications, and COP is the latest probability-based approach.
Note that they are only used as baselines to examine whether our method and other semi-supervised approaches obtain significantly improved performance.
\end{itemize}

\subsubsection{Parameter Settings and Implementations}

In RoSAS, the learning rate is set as 0.005, intermediate representation dimension $H$ is 128. As for the parameters in the loss function, we use $k=2$, $\alpha=0.5$, and $e=1$. Smooth-$\ell_1$ loss function is adopted as the base regression loss in ${L}$.
The batch size $b$ is 32. RoSAS uses the Adam optimizer with an $\ell_2$-norm weight decay regularizer. The temperature $T$ in dynamic weight averaging is $2$. RoSAS uses a multi-layer perceptron network structure since the used experimental datasets are multi-dimensional data. The representation module and the scoring module both adopt a one-hidden-layer structure. The number of hidden units in the representation network is set as $h_1 = D + \lfloor \frac{1}{2} (H-D) \rfloor$, and the scoring network uses $h_2 = \lfloor \frac{1}{2}H \rfloor $. We use LeakyReLU activation in the hidden layers and the $\tanh$ function to normalize final anomaly scores.

All the detectors are implemented in Python. The implementations of PReNet, DevNet, FeaWAD, DSAD, WSIF, and BiGAN are released by their original authors. The source code of TSS is publicly available. RoSAS, DSAD, and BiGAN employ the PyTorch framework, and PReNet, DevNet, and FeaWAD are based on Keras. We use implementations of COP and IF the \texttt{pyod} \citep{zhao2019pyod} package.

\subsubsection{Performance Evaluation Metrics and Computing Infrastructure
}
Following the popular experiment protocol of anomaly detection studies 
\citep{pang2019deep,pang2019prenet,ruff2020deep,xu2021beyond,xu2022dif}, two performance evaluation metrics, i.e., the Area under the Precision-Recall Curve (AUC-PR) and the Area under the Receiver Operating Characteristic Curve (AUC-ROC), are used. 
ROC curve indicates true positives against false positives, while points in PR curve are pairs of precision value and recall value of the anomaly class given different thresholds. 
These two metrics range from 0 to 1. Higher values indicate better performance. 
AUC-PR is more practical in real-world applications because it directly relates to benefits and costs of detection results, and achieving high AUC-PR is more challenging. 
Therefore, we take AUC-PR as the main detection performance metric in the following experiments. 
We report the average AUC-PR and AUC-ROC scores on each dataset over ten independent runs. 
Additionally, we employ the paired \textit{Wilcoxon} signed-rank test to determine if the AUC-ROC/AUC-PR of RoSAS and each of its contenders are significantly different. It can examine the statistical significance of the improvement of RoSAS against existing state-of-the-art performance. 

All the experiments are executed at a workstation with Intel Xeon Silver 4210R CPU, a single NVIDIA TITAN RTX GPU, and 64 GB RAM.

\subsection{Effectiveness}\label{sec:effectiveness}

\subsubsection{Anomaly Detection Performance on Real-world Datasets}

Following \citep{pang2019deep,pang2019prenet,wu2021surrogate,zhou2021feature}, we randomly select 30 true anomalies from the training data per dataset as anomaly examples and the remaining training data as the unlabeled set. 
RoSAS and its five contenders are trained on training sets and used to measure abnormal degrees of data samples in testing sets.
Labels of testing sets are strictly unknown to anomaly detectors and are only employed in the evaluation phase. 
As has been done in \citep{pang2019deep,pang2019prenet,zhou2021feature}, we also execute controlled experiments w.r.t. anomaly contamination rate. Each dataset is pre-processed by removing/injecting anomalies such that anomalies account for 2\% of the unlabeled set. Specifically, the injected anomaly examples are obtained by replacing the values of 5\% random features of a randomly selected real anomaly with the corresponding feature values of another real anomaly. This presents a simple and effective way to guarantee the presence of diverse and genuine (or weakly augmented) anomalies in the unlabeled data.
This pre-processing step can cancel out the influence of different contamination ratios such that the performance of these anomaly detectors is comparable across datasets from various domains. 
Please note that we also examine the performance w.r.t. a wide range of contamination ratios in the following experiment.

\begin{table*}[t]
\centering
\caption{AUC-PR and AUC-ROC performance ($\pm$ standard deviation) of RoSAS and its competing methods. The best performer is boldfaced.
}
\scalebox{0.7}{
	\begin{tabular}{
			p{0.13cm}| p{1.8cm} 
			p{1.7cm}<{\centering}p{1.7cm}<{\centering}
			p{1.7cm}<{\centering}p{1.7cm}<{\centering}
			p{1.7cm}<{\centering}p{1.7cm}<{\centering}
			p{1.7cm}<{\centering}p{1.7cm}<{\centering}
			p{0.9cm}<{\centering}p{0.9cm}<{\centering}
			p{0.9cm}<{\centering}
		}
		\toprule
		
		\multicolumn{1}{l}{}  & \multicolumn{1}{l}{\textbf{Data}} & 
		\textbf{RoSAS} & \textbf{PReNet} & \textbf{DevNet} & \textbf{FeaWAD} & \textbf{DSAD} & \textbf{TSS} & \textbf{WSIF} & \textbf{BiGAN} & \textbf{DIF} & \textbf{IF} & \textbf{COP} \\
		
		\midrule
		
		\multirow{13}[0]{*}{\rotatebox{90}{\textbf{AUC-PR}}} & DoH   & \textbf{0.893$_{\pm0.005}$} & 0.712$_{\pm0.021}$ & 0.628$_{\pm0.005}$ & 0.741$_{\pm0.068}$ & 0.842$_{\pm0.017}$ & 0.610$_{\pm0.002}$ & 0.546$_{\pm0.024}$ & 0.561$_{\pm0.055}$ & 0.396 & 0.427 & 0.340 \\
		& WebAttack & \textbf{0.781$_{\pm0.051}$} & 0.223$_{\pm0.055}$ & 0.220$_{\pm0.078}$ & 0.320$_{\pm0.081}$ & 0.458$_{\pm0.116}$ & 0.136$_{\pm0.005}$ & 0.026$_{\pm0.011}$ & 0.050$_{\pm0.034}$ & 0.003 & 0.004 & 0.004 \\
		& PortScan & \textbf{0.999$_{\pm0.000}$} & 0.983$_{\pm0.005}$ & 0.973$_{\pm0.030}$ & 0.995$_{\pm0.005}$ & 0.998$_{\pm0.001}$ & 0.990$_{\pm0.001}$ & 0.585$_{\pm0.131}$ & 0.605$_{\pm0.172}$ & 0.181 & 0.180 & 0.135 \\
		& Exploit & \textbf{0.740$_{\pm0.026}$} & 0.560$_{\pm0.056}$ & 0.450$_{\pm0.084}$ & 0.510$_{\pm0.075}$ & 0.623$_{\pm0.037}$ & 0.523$_{\pm0.076}$ & 0.199$_{\pm0.053}$ & 0.272$_{\pm0.081}$ & 0.255 & 0.060 & 0.083 \\
		& Backdoor & 0.877$_{\pm0.021}$ & 0.882$_{\pm0.005}$ & \textbf{0.884$_{\pm0.015}$} & 0.793$_{\pm0.095}$ & 0.666$_{\pm0.036}$ & 0.860$_{\pm0.018}$ & 0.437$_{\pm0.143}$ & 0.405$_{\pm0.152}$ & 0.394 & 0.052 & 0.069 \\
		& Thyroid & \textbf{0.839$_{\pm0.046}$} & 0.436$_{\pm0.024}$ & 0.252$_{\pm0.034}$ & 0.322$_{\pm0.038}$ & 0.304$_{\pm0.099}$ & 0.197$_{\pm0.010}$ & 0.537$_{\pm0.080}$ & 0.083$_{\pm0.006}$ & 0.074 & 0.131 & 0.134 \\
		& Donars & \textbf{1.000$_{\pm0.000}$} & 0.973$_{\pm0.015}$ & 0.999$_{\pm0.003}$ & 0.998$_{\pm0.005}$ & 0.999$_{\pm0.001}$ & 0.982$_{\pm0.007}$ & 0.780$_{\pm0.093}$ & 0.873$_{\pm0.075}$ & 0.115 & 0.238 & 0.242 \\
		& Fraud & \textbf{0.831$_{\pm0.003}$} & 0.803$_{\pm0.008}$ & 0.808$_{\pm0.004}$ & 0.596$_{\pm0.261}$ & 0.438$_{\pm0.109}$ & 0.800$_{\pm0.005}$ & 0.523$_{\pm0.073}$ & 0.664$_{\pm0.099}$ & 0.335 & 0.328 & 0.270 \\
		& Cover & \textbf{0.983$_{\pm0.003}$} & 0.957$_{\pm0.008}$ & 0.907$_{\pm0.055}$ & 0.939$_{\pm0.031}$ & 0.922$_{\pm0.026}$ & 0.916$_{\pm0.003}$ & 0.673$_{\pm0.154}$ & 0.603$_{\pm0.132}$ & 0.191 & 0.063 & 0.061 \\
		& Letter & \textbf{0.501$_{\pm0.026}$} & 0.332$_{\pm0.052}$ & 0.153$_{\pm0.077}$ & 0.270$_{\pm0.047}$ & 0.069$_{\pm0.027}$ & 0.163$_{\pm0.032}$ & 0.140$_{\pm0.043}$ & 0.067$_{\pm0.005}$ & 0.099 & 0.084 & 0.061 \\
		& Pendigits & \textbf{0.995$_{\pm0.013}$} & 0.906$_{\pm0.073}$ & 0.884$_{\pm0.053}$ & 0.907$_{\pm0.064}$ & 0.983$_{\pm0.023}$ & 0.916$_{\pm0.012}$ & 0.583$_{\pm0.117}$ & 0.161$_{\pm0.199}$ & 0.302 & 0.366 & 0.239 \\
		\cline{2-13}
		& \textit{Average} & \textbf{0.858$_{\pm0.018}$} & 0.706$_{\pm0.029}$ & 0.651$_{\pm0.040}$ & 0.672$_{\pm0.070}$ & 0.664$_{\pm0.045}$ & 0.645$_{\pm0.016}$ & 0.457$_{\pm0.084}$ & 0.395$_{\pm0.092}$ & 0.213 & 0.176 & 0.149 \\
		& \textit{p-value} & -     & 0.002  & 0.003  & 0.001  & 0.001  & 0.001  & 0.001  & 0.001  & 0.001  & 0.001  & 0.001  \\
		
		\midrule    
		
		\multirow{13}[0]{*}{\rotatebox{90}{\textbf{AUC-ROC}}} & DoH   & \multicolumn{1}{c}{\textbf{0.955$_{\pm0.001}$}} & 0.885$_{\pm0.004}$ & 0.884$_{\pm0.008}$ & 0.905$_{\pm0.017}$ & 0.932$_{\pm0.007}$ & 0.871$_{\pm0.001}$ & 0.753$_{\pm0.030}$ & 0.825$_{\pm0.056}$ & 0.693 & 0.674 & 0.676 \\
		& WebAttack & \multicolumn{1}{c}{0.988$_{\pm0.006}$} & 0.952$_{\pm0.004}$ & 0.934$_{\pm0.019}$ & 0.976$_{\pm0.012}$ & \textbf{0.989$_{\pm0.002}$} & 0.917$_{\pm0.001}$ & 0.830$_{\pm0.083}$ & 0.896$_{\pm0.012}$ & 0.530 & 0.607 & 0.630 \\
		& PortScan & \multicolumn{1}{c}{\textbf{0.999$_{\pm0.000}$}} & 0.997$_{\pm0.001}$ & 0.994$_{\pm0.006}$ & 0.999$_{\pm0.001}$ & 0.999$_{\pm0.000}$ & 0.997$_{\pm0.000}$ & 0.886$_{\pm0.079}$ & 0.910$_{\pm0.090}$ & 0.518 & 0.506 & 0.304 \\
		& Exploit & \multicolumn{1}{c}{\textbf{0.962$_{\pm0.010}$}} & 0.951$_{\pm0.006}$ & 0.913$_{\pm0.008}$ & 0.949$_{\pm0.018}$ & 0.956$_{\pm0.014}$ & 0.936$_{\pm0.008}$ & 0.803$_{\pm0.053}$ & 0.862$_{\pm0.021}$ & 0.864 & 0.743 & 0.771 \\
		& Backdoor & \multicolumn{1}{c}{\textbf{0.985$_{\pm0.004}$}} & 0.952$_{\pm0.002}$ & 0.971$_{\pm0.006}$ & 0.978$_{\pm0.004}$ & 0.965$_{\pm0.009}$ & 0.970$_{\pm0.005}$ & 0.816$_{\pm0.128}$ & 0.906$_{\pm0.018}$ & 0.915 & 0.753 & 0.791 \\
		& Thyroid & \multicolumn{1}{c}{\textbf{0.989$_{\pm0.003}$}} & 0.809$_{\pm0.010}$ & 0.728$_{\pm0.022}$ & 0.747$_{\pm0.011}$ & 0.729$_{\pm0.048}$ & 0.716$_{\pm0.018}$ & 0.904$_{\pm0.060}$ & 0.548$_{\pm0.020}$ & 0.497 & 0.646 & 0.703 \\
		& Donars & \multicolumn{1}{c}{\textbf{1.000$_{\pm0.000}$}} & 0.999$_{\pm0.000}$ & 1.000$_{\pm0.000}$ & 1.000$_{\pm0.000}$ & 1.000$_{\pm0.000}$ & 0.999$_{\pm0.000}$ & 0.986$_{\pm0.009}$ & 0.984$_{\pm0.014}$ & 0.780 & 0.891 & 0.846 \\
		& Fraud & \multicolumn{1}{c}{0.977$_{\pm0.003}$} & 0.967$_{\pm0.008}$ & 0.971$_{\pm0.002}$ & 0.973$_{\pm0.006}$ & 0.952$_{\pm0.015}$ & \textbf{0.979$_{\pm0.001}$} & 0.966$_{\pm0.004}$ & 0.933$_{\pm0.012}$ & 0.960 & 0.962 & 0.966 \\
		& Cover & \multicolumn{1}{c}{\textbf{1.000$_{\pm0.000}$}} & \textbf{1.000$_{\pm0.000}$} & 0.999$_{\pm0.001}$ & 0.999$_{\pm0.001}$ & 0.998$_{\pm0.002}$ & 0.999$_{\pm0.000}$ & 0.992$_{\pm0.005}$ & 0.975$_{\pm0.022}$ & 0.963 & 0.887 & 0.867 \\
		& Letter & \multicolumn{1}{c}{\textbf{0.873$_{\pm0.031}$}} & 0.822$_{\pm0.033}$ & 0.663$_{\pm0.106}$ & 0.770$_{\pm0.057}$ & 0.489$_{\pm0.078}$ & 0.707$_{\pm0.025}$ & 0.691$_{\pm0.033}$ & 0.546$_{\pm0.024}$ & 0.650 & 0.608 & 0.528 \\
		& Pendigits & \multicolumn{1}{c}{\textbf{1.000$_{\pm0.000}$}} & 0.999$_{\pm0.001}$ & 0.996$_{\pm0.001}$ & 0.999$_{\pm0.001}$ & \textbf{1.000$_{\pm0.000}$} & 0.997$_{\pm0.001}$ & 0.977$_{\pm0.011}$ & 0.705$_{\pm0.237}$ & 0.951 & 0.955 & 0.905 \\
		
		\cline{2-13}
		
		& \textit{Average} & \textbf{0.975$_{\pm0.005}$} & 0.939$_{\pm0.006}$ & 0.914$_{\pm0.016}$ & 0.936$_{\pm0.012}$ & 0.910$_{\pm0.016}$ & 0.917$_{\pm0.005}$ & 0.873$_{\pm0.045}$ & 0.826$_{\pm0.048}$ & 0.756 & 0.748 & 0.726 \\
		
		& \textit{p-value} &   -    & 0.005  & 0.005  & 0.008  & 0.017  & 0.007  & 0.001  & 0.001  & 0.001  & 0.001  & 0.001  \\
		
		\bottomrule
\end{tabular}}%
\label{tab:effectiveness}%
\end{table*}%

Table \ref{tab:effectiveness} shows the AUC-PR and the AUC-ROC performance of RoSAS and its competing methods. RoSAS achieves the best AUC-PR or AUC-ROC performance on all the datasets. According to the p-values in the \textit{Wilcoxon} signed-rank test, RoSAS significantly outperforms its ten competitors w.r.t. both AUC-PR and AUC-ROC at the 98\% confidence level. 
Averagely, RoSAS obtains a substantial performance leap (approximate 20\%-30\% AUC-PR improvement) over exiting state-of-the-art competing methods PReNet, DevNet, FeaWAD, DSAD, and TSS.
WSIF is a non-deep method, which is inferior to these deep state-of-the-art semi-supervised methods on complicated real-world datasets. 
BiGAN is originally designed for images, and its performance on tabular data might be downgraded. 
Benefiting from a few labeled anomalies, the average AUC-ROC performance of many semi-supervised methods exceeds 0.9, and RoSAS still gains 4\%-7\% improvement over state-of-the-art competitors. 
The performance of unsupervised anomaly detectors DIF, IF, and COP is distinctly inferior to all the semi-supervised approaches, which validates the importance of fully exploiting these readily accessible anomaly examples in real-world applications.

RoSAS achieves substantially superior detection performance with the help of the proposed contamination-resilient continuous supervision and feature learning-based regularization. 
The robustness of RoSAS is enhanced to better exploit the contaminated unlabeled set. RoSAS is also with direct fine-grained guidance to optimize anomaly scores more accurately. 
Therefore, RoSAS can better leverage dozens of anomaly examples and large-scale unlabeled data, resulting in effective semi-supervised anomaly detection. 
Note that PReNet obtains relatively better performance because it can resist small anomaly contamination thanks to its data combination operation. 
Other competitors do not consider the interference from noisy hidden anomalies and treat the whole unlabeled set as normal data. Also, all of these competing methods are only optimized by discrete supervision information that fails to indicate continuous abnormal degrees, resulting in suboptimal learning of anomaly scores.

\subsubsection{Capability of Handling Different Types of Anomalies}

\begin{figure}[t]
\centering	\includegraphics[width=0.9\textwidth]{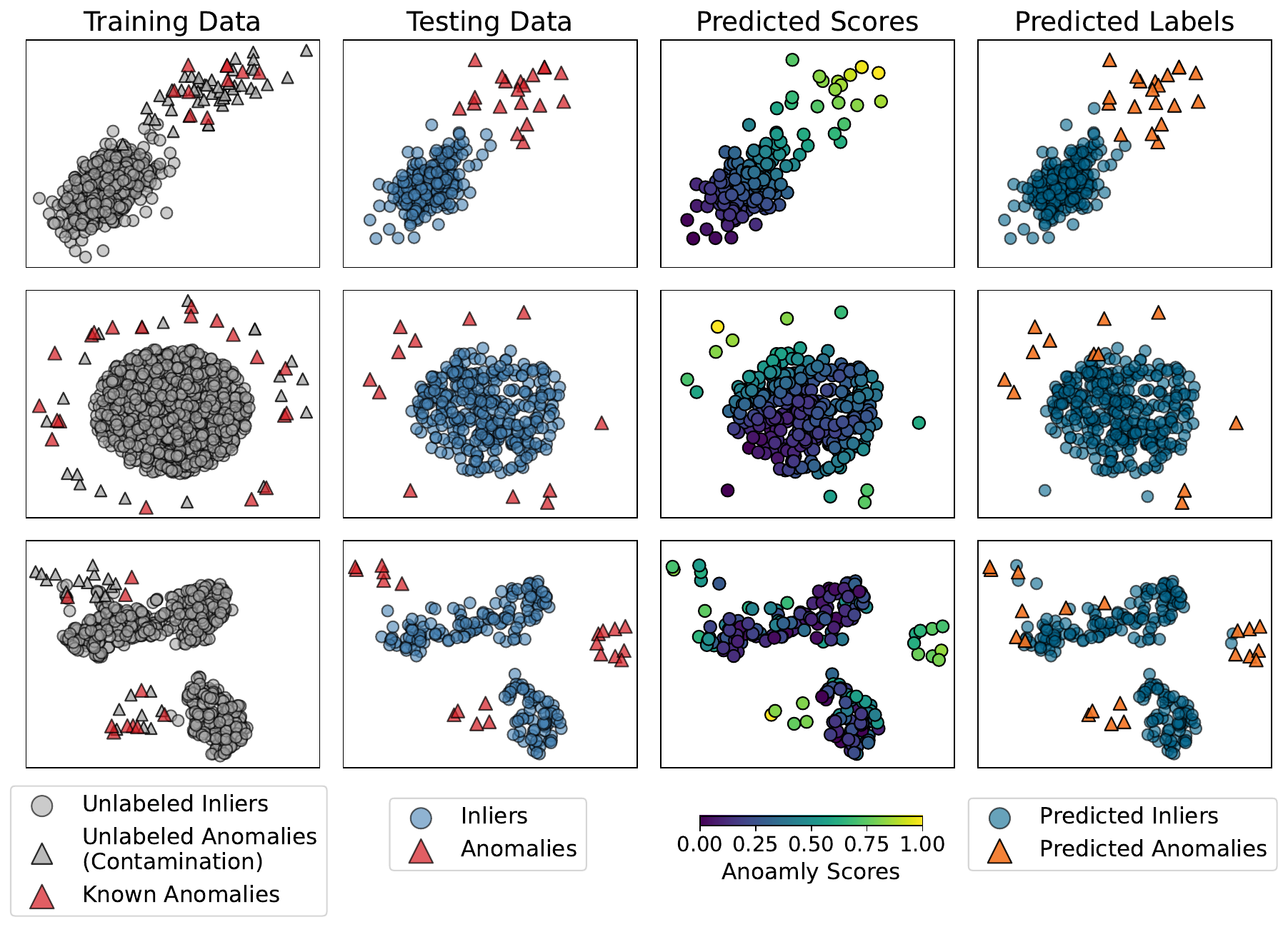}
\caption{
	Three toy cases with different anomaly types. Each row indicates a case. 
	The top and medium cases respectively contain \textit{clustered anomalies} and \textit{scattered anomalies}. The testing data of the bottom case has \textit{novel anomalies} (the anomaly cluster on the far right) that do not appear in the training set.  
	The panels in the left two columns show the data distribution of training/testing data, and the anomaly detection results of RoSAS including predicted anomaly scores and binary labels are visualized in the right two columns. 
}
\label{fig:case}
\end{figure}

We further investigate whether RoSAS can identify different types of anomalies.
Anomalies can be classified into \textit{clustered anomalies} and \textit{scattered anomalies} according to the intra-class proximity \citep{xu2019mix,zhou2022unseen}.
Clustered anomalies (e.g., diseases and fraudulent activities) share similar behaviors, while scattered anomalies (e.g., exceptions in industrial systems) randomly appear out of the inlier distribution and have weak or even no connections with other individual samples. 
Besides, in the semi-supervised setting, there might be some \textit{novel anomalies} that are different from labeled anomalies that appear during training. Novel anomalies are critical in real-world applications; for example, some advanced new attacks may pose severe threats to network security, but they are very different from those known intrusions.
Due to the difficulty of knowing specific anomaly types in real-world datasets, we create three synthetic cases to validate the capability of handling these anomaly types. 
Training and testing data distributions of these three cases are demonstrated in Figure \ref{fig:case}. Case 1 and Case 2 respectively contain clustered anomalies and scattered anomalies, and there is a cluster of novel anomalies in the testing set of Case 3.

Figure \ref{fig:case} further illustrates the detection results of RoSAS. 
By setting the threshold according to the size of true anomalies, we report both predicted anomaly scores and corresponding binary labels. We respectively analyze the detection results of the three cases below. 
\paragraph{Case 1}
In terms of clustered anomalies in Case 1, although the abnormal region is contaminated by unlabeled anomalies that are used as normal data, this region can be covered by new data samples that are augmented by our mass interpolation method. Labeled anomalies are over-sampled during training, and the unlabeled anomalies are still rare compared to genuine normal data because anomalies themselves are rare. The contamination can be corrected by new data generated via the interpolation of data combinations with correct labels, and thus RoSAS can effectively identify these clustered anomalies during inference.

\paragraph{Case 2}
As for scattered anomalies in Case 2, unlabeled anomalies may not largely influence the training process. However, one key issue of this case is the ``manifold intrusion'' problem. For instance, the interpolation into labeled anomalies may create augmented data samples in the normal distribution, but they are labeled by high abnormal degrees. To alleviate this problem, RoSAS uses Beta distribution with $\alpha\!=\!0.5$ in the mass interpolation process, thereby making most interpolation located in the local regions of original samples. 
This may still raise an inevitable limitation. Namely, RoSAS gives slightly higher anomaly scores to some margin points of the normal manifold, and there are two false positives as shown in the binary prediction results.

\paragraph{Case 3}
RoSAS is also applicable to identify novel anomalies that do not appear during training, as validated in Case 3. This advantage owes to the feature learning module of RoSAS. The learning objective posed upon the representation space judges whether labeled anomalies are effectively separated by introducing a reference divergence degree between unlabeled data. That is, this learning objective not only repels anomalies from the normal manifold but pulls unlabeled samples together. Therefore, during inference, novel anomalies can be far away from the normal manifold in the representation space.

\subsection{Robustness w.r.t. Anomaly Contamination Levels}\label{sec:robustness}

This experiment evaluates the robustness of RoSAS w.r.t. different anomaly contamination ratios (i.e., the proportion of anomalies in the unlabeled set $\mathcal{X}_U$). As anomalies are rare events in practical scenarios, we vary the contamination level from 0\% up to 8\%, and all the contamination levels use 30 random true anomalies as labeled data.

\begin{figure}[t]
\centering 
\subfigcapskip=-5pt
\subfigure[]{
	\includegraphics[width=0.99\linewidth]{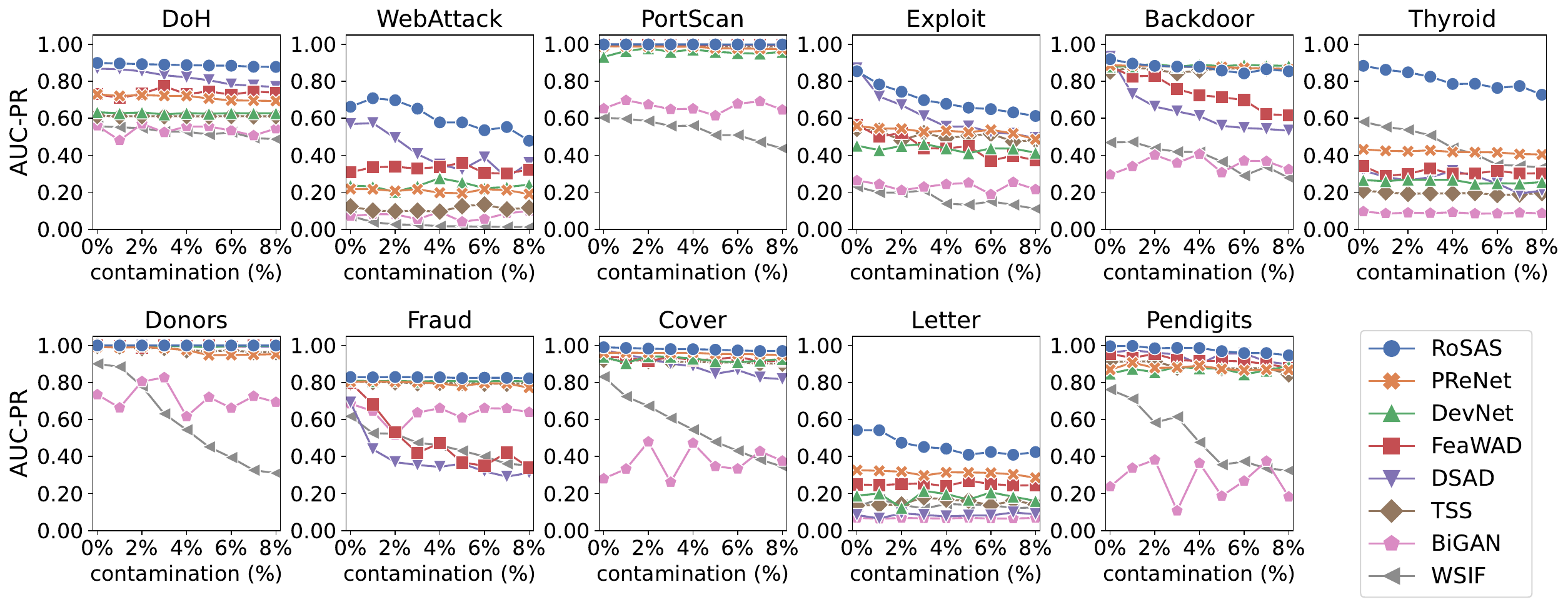}}
\subfigure[]{
	\includegraphics[width=0.99\linewidth]{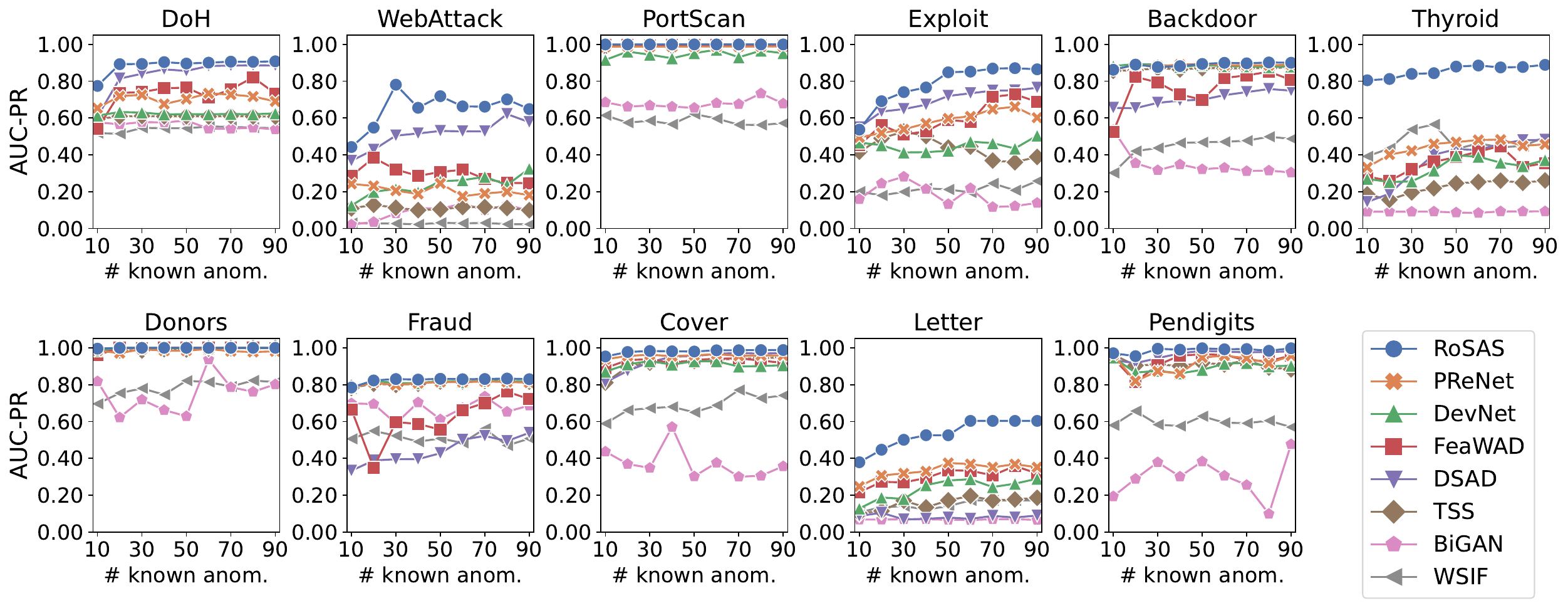}}
\caption{AUC-PR results of RoSAS and its semi-supervised competing methods on datasets with \textbf{(a)} different anomaly contamination levels (i.e., ratios of anomalies in the unlabeled set $\mathcal{X}_U$) and \textbf{(b)} varying numbers of labeled anomalies (i.e., the size of labeled anomaly examples $\mathcal{X}_A$). }
\label{fig:robust_efficacy}
\end{figure}

Figure \ref{fig:robust_efficacy} (a) shows the AUC-PR results on all the eleven real-world datasets with varying anomaly contamination levels. Anomaly detection performance generally decreases when the contamination level increases. Nevertheless, in the vast majority of cases, RoSAS is more robust than the competitors. 
It is noteworthy that, in some datasets (e.g., \textit{DoH}, \textit{PortScan}, and \textit{Pendigits}), most anomaly detectors are stable when increasing the contamination rate. It might be because the increased anomalies are isolated data samples, and anomaly detection models can easily filter the interference. The competitors can also obtain very competitive performance on these datasets.
However, in complicated datasets like \textit{Exploit}, \textit{Backdoor}, \textit{Thyroid}, and \textit{Fraud}, these anomalies that are hidden in the unlabeled set greatly blur the boundary between normal data and anomalies. RoSAS can consistently obtain better performance than the competitors in challenging noisy environments with high contamination levels. 


\subsection{Data Efficacy of Labeled Anomalies}\label{sec:efficacy}

This experiment estimates the data efficacy of different numbers of labeled anomalies in terms of the value they bring to semi-supervised anomaly detection. 
In other words, this experiment examines whether RoSAS achieves more significant performance improvement than its competing methods when more labeled anomalies are available. The number of labeled anomalies is increased from 10 to 90, and the contamination level is maintained at 2\%.

Figure \ref{fig:robust_efficacy} (b) shows the AUC-PR results of RoSAS and its contenders w.r.t varying numbers of labeled anomalies. Semi-supervised anomaly detectors generally perform better when more labeled anomalies are accessible. However, this law is not always true in practice. Some anomaly detectors also present fluctuation trends on some datasets. These increased labeled anomalies may have heterogeneous behaviors and carry conflicting information which imposes negative effects on anomaly detectors, as has been explained in \citep{pang2019deep}. 
By contrast, our method obtains more stable and superior performance by fully utilizing limited labeled anomalies. It is noteworthy that some detectors also do not perform better when more labeled anomalies are available. It might be because these increased labeled anomalies have very similar behaviors and fail to bring useful information related to the anomaly distribution.

\subsection{Scalability Test} \label{sec:scalability}

\begin{figure}[t]
	\centering	\includegraphics[width=0.8\textwidth]{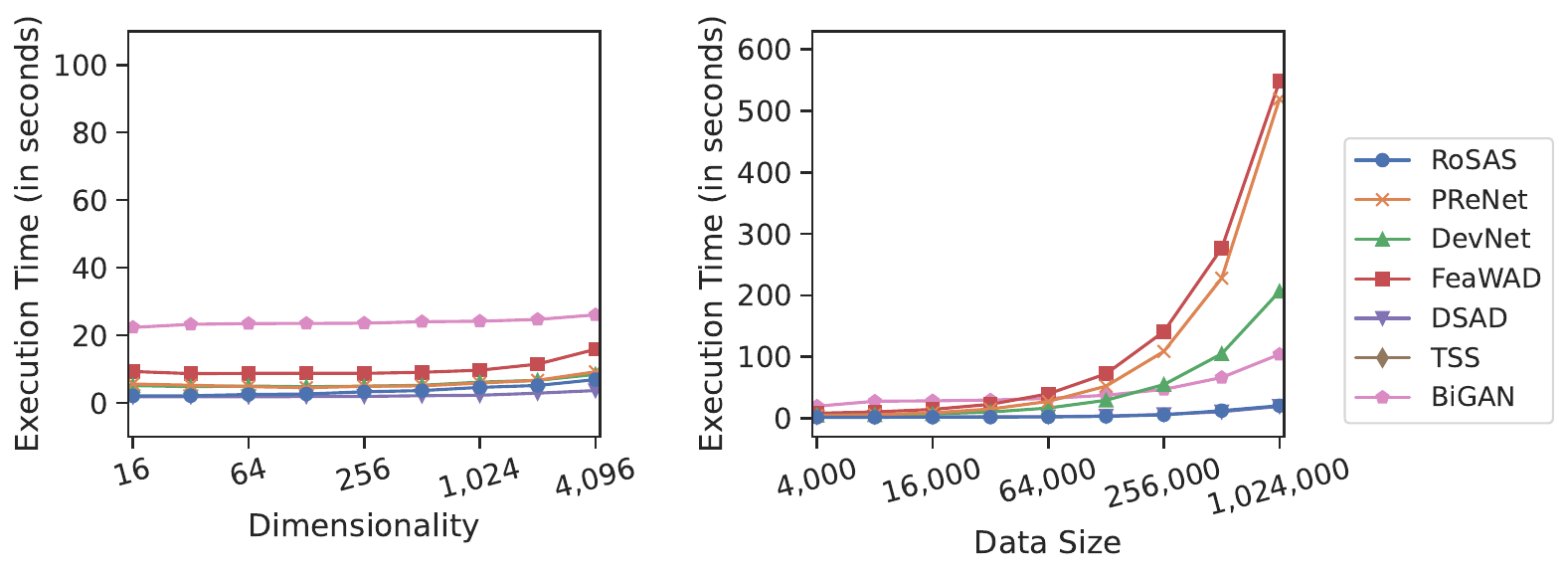}
	\caption{Scalability test results. 
	} 
	\label{fig:scalabiliy}
\end{figure}

This experiment evaluates the scalability of RoSAS.
Nine datasets are created with the same data size (i.e., 5,000) and dimensionality increasing in multiples of 2 from 16 to 4,096. 
Another nine datasets are generated with varying data sizes increasing from 4,000 to 1,024,000 with fixed dimensions (i.e., 128). 
For the sake of comparison fairness, we employ deep semi-supervised anomaly detection methods (i.e., PReNet, DevNet, FeaWAD, DSAD, TSS, and BiGAN) as counterparts in this experiment. 
We use the same training configuration for these methods including the size of mini-batches (32) and the number of mini-batches per training epoch (20).
We report the execution time including the training time of 10 epochs and the inference time.
Scalability test results are reported in Figure \ref{fig:scalabiliy}. RoSAS and its counterparts can efficiently handle high-dimensional data thanks to the parallel accelerators in the mini-batch calculation of GPU. 
RoSAS only takes less than 10 seconds when handling 4,096-dimensional data. 
In terms of the scale-up test w.r.t. data size, RoSAS, DSAD, and TSS have comparably good efficiency. In comparison, FeaWAD and PReNet have complicated network structures that lead to significantly increased execution time. RoSAS uses about 20 seconds to handle the dataset containing 1,024,000 data samples.

\begin{table}[t]
	\centering
	\caption{AUC-PR results of RoSAS and its five ablated versions with the improvement rates of RoSAS compared to its variants per dataset. Positive rates are boldfaced.    }
	\scalebox{0.7}{
		\begin{tabular}{lllllll}
			
			\toprule
			\textbf{Data}  
			& \textbf{RoSAS} 
			& \boldmath{}$L$$\rightarrow$$L_{dis}$\unboldmath{}
			& \boldmath{}$L$$\rightarrow$$L_{dev}$\unboldmath{}
			& \boldmath{}$L$$\rightarrow$$L_{reg}$\unboldmath{}
			& \boldmath{}\textbf{w/o} $L'$\unboldmath{} 
			& \boldmath{}\textbf{w/o} $L_c$\unboldmath{} \\
			\midrule
			DoH   & 0.893$_{\pm0.005}$ & 0.894$_{\pm0.006}$ ({-0.1\%}) & 0.888$_{\pm0.011}$ (\textbf{0.6\%}) & 0.867$_{\pm0.008}$ (\textbf{3.0\%}) & 0.892$_{\pm0.006}$ (\textbf{0.1\%}) & 0.891$_{\pm0.005}$ (\textbf{0.2\%}) \\
			WebAttack & 0.781$_{\pm0.051}$ & 0.623$_{\pm0.146}$ (\textbf{25.4\%}) & 0.508$_{\pm0.136}$ (\textbf{53.7\%}) & 0.434$_{\pm0.090}$ (\textbf{80.0\%}) & 0.542$_{\pm0.081}$ (\textbf{44.1\%}) & 0.638$_{\pm0.139}$ (\textbf{22.4\%}) \\
			PortScan & 0.999$_{\pm0.000}$ & 0.999$_{\pm0.000}$ ({0.0\%}) & 0.999$_{\pm0.000}$ ({0.0\%}) & 0.997$_{\pm0.001}$ (\textbf{0.2\%}) & 0.997$_{\pm0.001}$ (\textbf{0.2\%}) & 0.999$_{\pm0.000}$ ({0.0\%}) \\
			Exploit & 0.740$_{\pm0.026}$ & 0.727$_{\pm0.031}$ (\textbf{1.8\%}) & 0.674$_{\pm0.038}$ (\textbf{9.8\%}) & 0.632$_{\pm0.069}$ (\textbf{17.1\%}) & 0.694$_{\pm0.031}$ (\textbf{6.6\%}) & 0.721$_{\pm0.034}$ (\textbf{2.6\%}) \\
			Backdoor & 0.877$_{\pm0.021}$ & 0.873$_{\pm0.020}$ (\textbf{0.5\%}) & 0.858$_{\pm0.024}$ (\textbf{2.2\%}) & 0.888$_{\pm0.011}$ ({-1.2\%}) & 0.892$_{\pm0.009}$ ({-1.7\%}) & 0.877$_{\pm0.020}$ ({0.0\%}) \\
			Thyroid & 0.839$_{\pm0.046}$ & 0.623$_{\pm0.167}$ (\textbf{34.7\%}) & 0.543$_{\pm0.190}$ (\textbf{54.5\%}) & 0.777$_{\pm0.021}$ (\textbf{8.0\%}) & 0.812$_{\pm0.049}$ (\textbf{3.3\%}) & 0.833$_{\pm0.044}$ (\textbf{0.7\%}) \\
			Donars & 1.000$_{\pm0.000}$ & 1.000$_{\pm0.000}$ ({0.0\%}) & 1.000$_{\pm0.000}$ ({0.0\%}) & 1.000$_{\pm0.000}$ ({0.0\%}) & 1.000$_{\pm0.000}$ ({0.0\%}) & 1.000$_{\pm0.000}$ ({0.0\%}) \\
			Fraud & 0.831$_{\pm0.003}$ & 0.830$_{\pm0.003}$ (\textbf{0.1\%}) & 0.824$_{\pm0.008}$ (\textbf{0.8\%}) & 0.820$_{\pm0.005}$ (\textbf{1.3\%}) & 0.819$_{\pm0.005}$ (\textbf{1.5\%}) & 0.830$_{\pm0.004}$ (\textbf{0.1\%}) \\
			Cover & 0.983$_{\pm0.003}$ & 0.980$_{\pm0.005}$ (\textbf{0.3\%}) & 0.982$_{\pm0.004}$ (\textbf{0.1\%}) & 0.973$_{\pm0.005}$ (\textbf{1.0\%}) & 0.985$_{\pm0.002}$ ({-0.2\%}) & 0.983$_{\pm0.004}$ ({0.0\%}) \\
			Letter & 0.501$_{\pm0.026}$ & 0.433$_{\pm0.059}$ (\textbf{15.7\%}) & 0.417$_{\pm0.039}$ (\textbf{20.1\%}) & 0.333$_{\pm0.049}$ (\textbf{50.5\%}) & 0.372$_{\pm0.050}$ (\textbf{34.7\%}) & 0.494$_{\pm0.064}$ (\textbf{1.4\%}) \\
			Pendigits & 0.995$_{\pm0.013}$ & 0.990$_{\pm0.014}$ (\textbf{0.5\%}) & 0.987$_{\pm0.015}$ (\textbf{0.8\%}) & 0.978$_{\pm0.011}$ (\textbf{1.7\%}) & 0.999$_{\pm0.001}$ ({-0.4\%}) & 0.994$_{\pm0.012}$ (\textbf{0.1\%}) \\
			
			\hline
			\textit{Average} & 0.858$_{\pm0.018}$ & 0.816$_{\pm0.041}$ (\textbf{7.1\%}) & 0.789$_{\pm0.042}$ (\textbf{13.0\%}) & 0.791$_{\pm0.025}$ (\textbf{14.7\%}) & 0.819$_{\pm0.021}$ (\textbf{8.0\%}) & 0.842$_{\pm0.030}$ (\textbf{2.5\%}) \\
			
			\textit{p-value} &   -    
			& \multicolumn{1}{l}{0.013}  
			& \multicolumn{1}{l}{0.008}  & \multicolumn{1}{l}{0.014}
			& \multicolumn{1}{l}{0.126}  & \multicolumn{1}{l}{0.018}  \\
			
			\bottomrule
		\end{tabular}%
	}
	\label{tab:ablation}%
\end{table}%

\subsection{Ablation Study}\label{sec:ablation}
This experiment is to validate the contribution of key designs in RoSAS. 
We set five ablated versions. The changes in these variants are introduced as follows, and other parts are the same as RoSAS. 
\begin{itemize}
	\item \boldmath{}$L$$\rightarrow$$L_{dis}$\unboldmath{} discretizes the generated continuous supervision targets used in the anomaly scoring loss function $L$.
	\item \boldmath{}$L$$\rightarrow$$L_{dev}$\unboldmath{} uses state-of-the-art anomaly scoring loss function used in DevNet \citep{pang2019deep,pang2021explainable} to replace $L$.
	\item \boldmath{}$L$$\rightarrow$$L_{reg}$\unboldmath{} uses a bare regression loss function used in RoSAS (i.e., smooth-$\ell_1$ loss) to replace $L$. 
	\item \textbf{w/o} \boldmath{}$L'$\unboldmath{} removes the feature learning-based regularizer $L'$. 
	\item \textbf{w/o} \boldmath{}$L_{c}$\unboldmath{} removes the consistency learning part in $L$.
\end{itemize}

The first three variants (\boldmath{}$L$$\rightarrow$$L_{dis}$\unboldmath{}, \boldmath{}$L$$\rightarrow$$L_{dev}$\unboldmath{}, and \boldmath{}$L$$\rightarrow$$L_{reg}$\unboldmath{}) only take \textit{discrete supervision information} to optimize anomaly scores, which are used to verify the significance of continuous supervision-guided anomaly score optimization.
The ablated variants \textbf{w/o} \boldmath{}$L'$\unboldmath{} and \textbf{w/o} \boldmath{}$L_{c}$\unboldmath{} measure the contributions of the feature learning-based regularization $L'$ and the consistency constraint in $L$.

The AUC-PR performance of RoSAS and its five ablated variants is shown in Table \ref{tab:ablation}. RoSAS significantly outperforms its three ablated versions \boldmath{}$L$$\rightarrow$$L_{dis}$\unboldmath{}, \boldmath{}$L$$\rightarrow$$L_{dev}$\unboldmath{}, and \boldmath{}$L$$\rightarrow$$L_{reg}$\unboldmath{} at 98\% confidence level. More than 7\% average improvement rate is achieved. These three variants use various objectives with only discrete supervision information.
This comparison result validates the significance of using continuous supervision signals in anomaly score optimization. Anomalies are with various abnormal degrees, and anomaly scores naturally take on a continuous distribution. It is hard for discrete supervision information to accurately describe such consecutive trends in continuous distribution, resulting in suboptimal optimization of anomaly scores in these real-world datasets.  
Our work reveals this significant limitation in current anomaly detection studies and devises a simple but effective solution.
On the other hand, RoSAS outperforms \textbf{w/o} \boldmath{}$L'$\unboldmath{} by 8\% and at 85\% confidence level, which verifies the complementary robustness enhancement effect brought by the feature learning-based regularization. Compared to \textbf{w/o} \boldmath{}$L_c$\unboldmath{}, the average improvement is above 2\% and the confidence interval is 98\%, which quantitatively measures the contribution of the consistency learning in producing smoother anomaly scores.

\subsection{Sensitivity Test}\label{sec:sensitivity}
\begin{figure}[t]
	\centering	\includegraphics[width=0.8\textwidth]{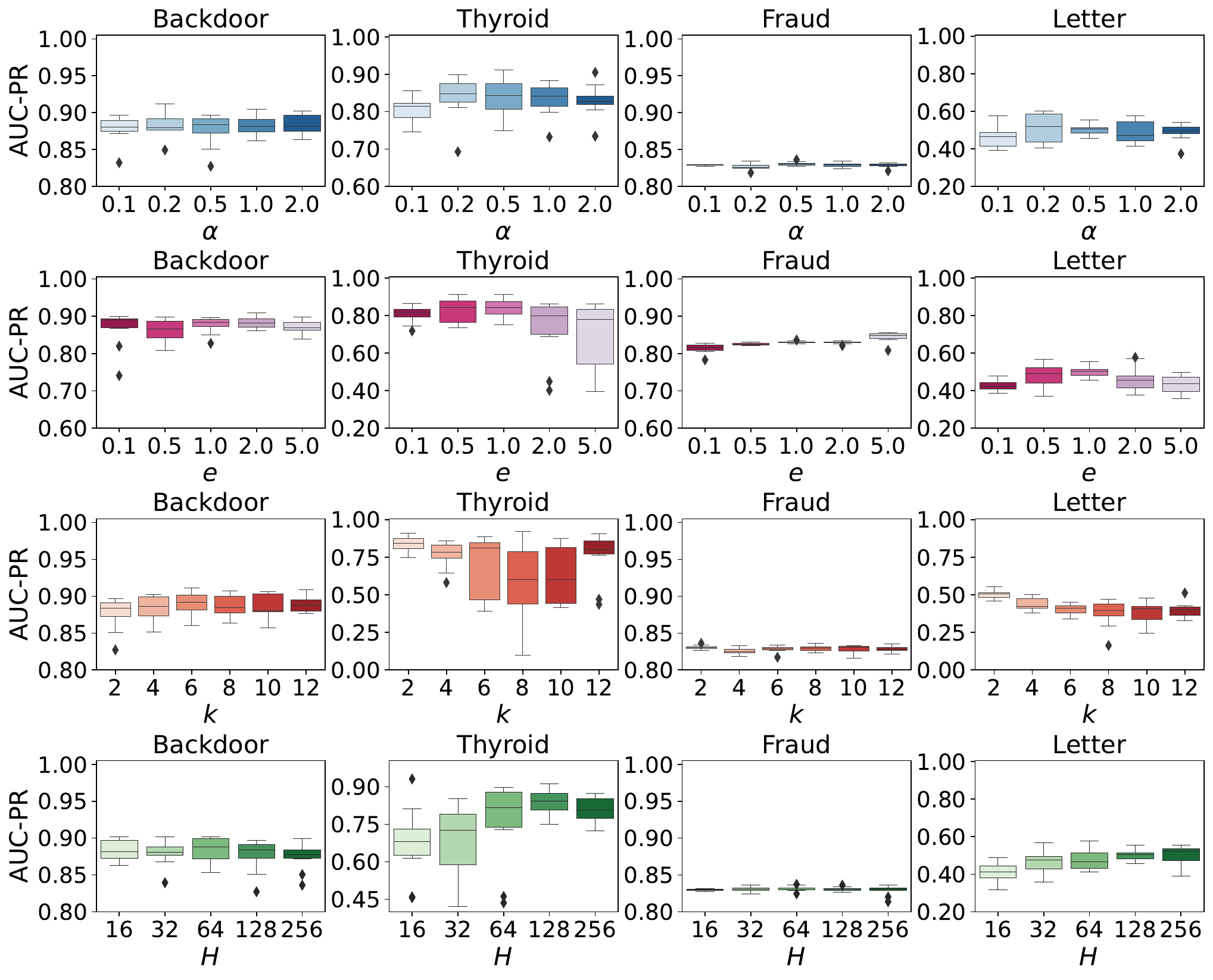}
	\caption{AUC-PR results of RoSAS with different settings of four key hyper-parameters ($\alpha$, $e$, $k$, and $H$).
	} 
	\label{fig:sens}
\end{figure}

We investigate the influence of different settings of key hyper-parameters in RoSAS, i.e., $\alpha$ in the Beta distribution, $k$ in the mass interpolation, $e$ in the regularizer, and the intermediate representation dimension $H$. These hyper-parameters are tuned in turn and other parameters are kept the same as previously reported. RoSAS is performed 10 times on each hyper-parameter setting. 
The box plot of 10 AUC-PR values per dataset is illustrated in Figure \ref{fig:sens}. 
We show four representative datasets, and the other seven datasets are with similar or stable trends.  
As analyzed before, $\alpha$ may influence the detection performance to some extent, and we use $\alpha\!=\!0.5$ considering the ``manifold intrusion'' problem. 
Besides, we can safely use a margin $e\!=\!1$ in the feature learning-based regularizer. 
The choice of $k$ might considerably influence the detection performance, and $k=2$ is more stable. Lower representation dimension $H$ fails to convey sufficient information to the downstream anomaly scoring process, and thus 128 is recommended.

\section{Discussion and Implications}

\subsection{Key contributions}

This study first summarizes the prior arts of this research field by giving a general semi-supervised anomaly detection framework. This framework contributes a unifying view of this research line, and we theoretically show how representative existing methods are instantiated from this framework. 
It may also offer valuable insights into the design of new semi-supervised anomaly detection models. 
More importantly, we uncover the key limitations of supervisory signals directly supplied by the semi-supervised setting and broadly used in existing methods. Motivated by these problems, we further propose a concrete anomaly detection method, and specifically, we make the following technical contributions.

This study contributes to the semi-supervised anomaly detection literature by taking into account the anomaly contamination problem. Arguably, many prior works directly use the whole unlabeled set as normal data for training their models (see Table \ref{tab:framework} and Section \ref{sec:limitation}), and their performance is considerably downgraded by these noisy points (as illustrated in the toy case in Figure \ref{fig:example} and real-world datasets in Table \ref{tab:effectiveness}). Our method RoSAS is shown to be a simple yet effective solution to address this limitation. Instead of directly feeding original flawed supervision into the learning model, we propose new supervision containing augmented data with more reliable label information, resulting in stronger robustness than existing state-of-the-art methods when the training set is with high contamination level (see empirical results in Figure \ref{fig:robust_efficacy}).

We consider our work as a starting point for leveraging continuous supervision information to optimize continuously distributed anomaly scores. To the best of our knowledge, we are the first to raise this issue in anomaly detection. We empirically show the advantage of using continuous supervision over discretized ones (see Table \ref{tab:ablation}). Continuous supervision can lead to significant performance gain at 98\% confidence level. Also, we pose a consistency constraint to further enhance the capability of producing smoother anomaly scores, which brings about 5\% performance improvement. These findings may foster future theoretical research or inspire new optimization mechanisms of anomaly scores.

To sum up, different from current studies that rely on \textit{contaminated discrete supervision}, our core novelty is a new kind of \textit{contamination-resilient continuous supervision}. This supervision better conforms to the real abnormal-normal distribution and offers significantly better guidance to the optimization of the end-to-end anomaly scoring neural network.

\subsection{Practical Implications}

Albeit a plethora of unsupervised anomaly detection models, many real-world systems are looking for anomaly detectors that can exploit their historical anomalies, and this study adds a new competitive option to the list that currently only contains limited choices.  
We show that only 30 anomalies can bring drastically improved performance than unsupervised models that work on unlabeled data only (e.g., our approach RoSAS achieves 0.999 of AUC-PR on an intrusion detection dataset \textit{PortScan} while unsupervised performance is only as low as 0.1). 
Given such huge benefits, instead of digging into the design of unsupervised anomaly detection models, one quick way to boost detection performance might be to transfer the unsupervised setting to the semi-supervised paradigm by feeding a few anomaly examples.

There are also many research and development fronts that we are pursuing in the future to further enhance the practical impact of this research. 
On one hand, this study can be extended to applications in different fields. We employ eleven datasets mainly from three domains including cybersecurity, medicine, and finance, and our approach also has the potential to identify system faults in AIOps or attacks in AI safety. On the other hand, by plugging in advanced network structures, our approach can be also applied to handle different data types (e.g., Transformer for sequential data, graph neural networks for graph data, and convolutional networks for images).


\section{Conclusions}
This paper first presents a general framework of deep semi-supervised anomaly detection to summarize this research line and reveal two key limitations of current studies.  
We then propose RoSAS, a concrete deep semi-supervised anomaly detection method.
By optimizing the detection model using the mass-interpolation-based continuous supervision that explicitly indicates faithful abnormal degrees, RoSAS learns accurate and noise-tolerate anomaly scores. Through extensive empirical results, we show two key advantages of using our continuous supervisory signals compared to the current discrete one: 1) our approach is substantially more robust w.r.t. anomaly contamination, especially on challenging cases with high contamination levels; 2) it is more data-efficient, that is, different numbers of labeled anomalies can be fully leveraged. These advantages are the main drivers of the overall superior performance of RoSAS that achieves about 20\%-30\% AUC-PR improvement over state-of-the-art semi-supervised anomaly detection approaches on 11 real-world datasets. 


\section*{Acknowledgments}

Hongzuo Xu, Yijie Wang, Songlei Jian, Ning Liu, and Yongjun Wang are supported in part by the National Key R\&D Program of China under Grant 2022ZD0115302, in part by the National Natural Science Foundation of China under Grants 62002371 and 61379052, in part by the Science Foundation of Ministry of Education of China under Grant 2018A02002, in part by the Postgraduate Scientific Research Innovation Project of Hunan Province under Grants CX20210049 and CX20210028, in part by the Natural Science Foundation for Distinguished Young Scholars of Hunan Province under Grant 14JJ1026, and the Foundation of National University of Defense Technology under Grant ZK21-17. Guansong Pang is supported in part by the Singapore Ministry of Education (MOE) Academic Research Fund (AcRF) Tier 1 under Grant 21SISSMU031.

We also thank the referees for their comments, which helped improve this paper considerably.

\bibliographystyle{model5-names}
\bibliography{main-r1}

\end{document}